# Benchmarking Next-Generation Reasoning-Focused Large Language Models in Ophthalmology: A Head-to-Head Evaluation on 5,888 Items

Minjie Zou[1,2]*, Sahana Srinivasan[1,2]*, Thaddaeus Wai Soon Lo[3]*, Ke Zou[1,2], Gabriel Dawei Yang[3,4], Xuguang Ai[5], Hyunjae Kim[5], Maxwell Singer[6], Fares Antaki[7,8], Kelvin Li[9], Robert Chang[10], Marcus Tan[1,2,11], David Ziyou Chen[1,2,11], Dianbo Liu[1,2], Qingyu Chen[5]†, Yih Chung Tham[1,2,3,12]†

1. Centre for Innovation and Precision Eye Health, Yong Loo Lin School of Medicine, National University of Singapore, Singapore.
2. Department of Ophthalmology, Yong Loo Lin School of Medicine, National University of Singapore, Singapore.
3. Singapore Eye Research Institute, Singapore National Eye Centre, Singapore
4. Department of Ophthalmology and Visual Sciences, The Chinese University of Hong Kong, Hong Kong Special Administrative Region, China
5. Department of Biomedical Informatics and Data Science, Yale School of Medicine, Yale University, New Haven, USA
6. Department of Ophthalmology and Visual Science, Yale School of Medicine, Yale University, New Haven, USA
7. Cole Eye Institute, Cleveland Clinic, Cleveland, OH, USA
8. The CHUM School of Artificial Intelligence in Healthcare, Montreal, QC, Canada
9. Department of Ophthalmology, Tan Tock Seng Hospital, National Healthcare Group, Singapore
10. Department of Ophthalmology, Byers Eye Institute, Stanford University, Stanford, California, USA

11. Department of Ophthalmology, National University Hospital, Singapore.

12. Eye Academic Clinical Program (Eye ACP), Duke NUS Medical School, Singapore.

*Contributed equally as first authors

†Contributed equally as last authors

Recent advances in reasoning-focused large language models (LLMs) mark a shift from general LLMs toward models designed for complex decision-making, a crucial aspect in medicine. However, their performance in specialized domains like ophthalmology remains underexplored. This study comprehensively evaluated and compared the accuracy and reasoning capabilities of four newly developed reasoning-focused LLMs, namely DeepSeek-R1, OpenAI o1, o3-mini, and Gemini 2.0 Flash-Thinking. Each model was assessed using 5,888 multiple-choice ophthalmology exam questions from the MedMCQA dataset in zero-shot setting. Quantitative evaluation included accuracy, Macro-F1, and five text-generation metrics (ROUGE-L, METEOR, BERTScore, BARTScore, and AlignScore), computed against ground-truth reasonings. Average inference time was recorded for a subset of 100 randomly selected questions. Additionally, two board-certified ophthalmologists qualitatively assessed clarity, completeness, and reasoning structure of responses to differential diagnosis questions.O1 (0.902) and DeepSeek-R1 (0.888) achieved the highest accuracy, with o1 also leading in Macro-F1 (0.900). The performance of models across the text-generation metrics varied: O3-mini excelled in ROUGE-L (0.151), o1 in METEOR (0.232), DeepSeek-R1 and o3-mini tied for BERTScore (0.673), DeepSeek-R1 (-4.105) and Gemini 2.0 Flash-Thinking (-4.127) performed best in BARTScore, while o3-mini (0.181) and o1 (0.176) led AlignScore. Inference time across the models varied, with DeepSeek-R1 being slowest (40.4 seconds) and Gemini 2.0 Flash-Thinking fastest (6.7 seconds). Qualitative evaluation revealed that DeepSeek-R1 and Gemini 2.0 Flash-Thinking tended to provide detailed and comprehensive intermediate reasoning, whereas o1 and o3-mini displayed concise and summarized justifications.

**Introduction**

Large language models (LLMs) are increasingly shaping the healthcare sector, with potential applications across various clinical settings, enhancing accessibility and health service delivery.[1-10] This field has recently entered a new phase with the emergence of reasoning-focused LLMs for improved logical consistency, interpretability and credibility, beginning with OpenAI's o1 LLM that was released in September 2024.[11] This shift continued with the introduction of DeepSeek-R1, an open-weighted LLM designed with emphasis on reasoning capabilities, incorporating techniques such as multi-headed latent attention and a mixture-of-experts (MoE) architecture to improve training speed and inference efficiency.[12-14] Unlike previous LLMs (e.g. ChatGPT-4, 4o, Gemini 1.5 etc.) that require explicit chain-of-thought (CoT) prompts, these new reasoning-focused LLMs inherently structure responses with clear logical progression, potentially improving transparency and interpretability for users.[15,16]

Recently, OpenAI's o3-mini and Google DeepMind's Gemini 2.0 Flash-Thinking have also been released, both comprising enhanced reasoning capabilities.[17-19] OpenAI claimed that o3-mini has comparable performance to o1 in general reasoning tasks but with higher speed and lower costs[17]. On the other hand, Gemini 2.0 Flash-Thinking could generate responses to complex tasks without high latencies, but it remains an experimental model lacking in built-in tool usage such as web search capabilities.[18,19] These recent developments underscore the rapid evolution of LLMs towards more structured and sophisticated reasoning, with potential to support decision making in medicine.

The capabilities of these reasoning-focused LLMs extend beyond simple information retrieval, and enable LLMs to generate nuanced, context-aware responses while following a CoT reasoning style.[20-24] Such reasoning features could be particularly useful in healthcare settings, where reasoning-focused LLMs may help to break down complex tasks into smaller, structured reasoning steps, offering clearer insights for both patients and physicians.[15] However, despite recent claims of advanced reasoning capabilities for DeepSeek-R1 and the high accuracy across general tasks,[15,25-27] whether these enhancements truly translate into superior performance within specialized medical domains remains largely unexamined. Given the demand in ophthalmology for precise diagnostic reasoning and nuanced clinical decision-making, it serves as an ideal 'medical domain testbed' for evaluating the performances of these next-generation reasoning-focused LLMs.[21]

In this study, using a large ophthalmological exam benchmark dataset of 5,888 items, we aimed to comprehensively evaluate and compare the performance of four recently released reasoning-focused LLMs – DeepSeek-R1, OpenAI o1, OpenAI o3-mini, and Gemini 2.0 Flash-Thinking. This benchmark involved assessing their accuracy, reasoning capabilities, and processing speed. Findings from this study would provide unique insights into the potential applicability and limitations of these reasoning-focused LLMs, giving deeper understanding of their potential for clinical decision support.

**Methods**

**Models Evaluated**

An overview of this study is presented in Figure 1. This study included four reasoning-focused LLMs, namely DeepSeek-R1, OpenAI o1 (o1-preview-2024-09-12), o3-mini (o3-mini-2025-01-31), and Gemini 2.0 Flash-Thinking. The key characteristics of these LLMs are summarized in **Table 1**. In brief, these models vary in architecture, reasoning capabilities, and extent of transparency. For instance, DeepSeek-R1 (released January 2025) employs an MoE design and displays more thorough reasoning processes.[28] OpenAI o1 (released September 2024) is known for its enhanced "chain-of-thought" reasoning, though its architectural details remain undisclosed.[11] Its sister model, o3-mini, was released in January 2025, and is a more efficient variant of o1 in terms of computational resources and financial costs.[17] Finally, Gemini 2.0 Flash-Thinking (December 2025) uses a dropdown style step-by-step reasoning approach, highlighting the explainability of its responses. **Figure 2** presents user interface screenshots for each of the four reasoning-focused LLMs.

The comparative assessment was conducted using the ophthalmology subset of the MedMCQA dataset, employing both accuracy- and reasoning-focused metrics (details below). OpenAI's o1, o3-mini, and Gemini 2.0 Flash-Thinking were accessed through their respective proprietary servers and APIs. DeepSeek-R1 was accessed through the Microsoft Azure AI Foundry.

**Dataset**

**Supplementary Figure 1** illustrates the dataset curation process of this study. We utilized the ophthalmology subset of the MedMCQA dataset, which comprised of 6,990 multiple-choice question (MCQ) items as previously described.[29] Each MCQ item consisted of a question and four accompanying options, of which only one option was correct. Additionally, each MCQ item contained an explanation of the solution, which served as the ground truth reasoning to evaluate the outputs generated by the four LLMs in this study.

To ensure that the ground truth reasonings were of a high quality, we utilized OpenAI GPT-4o-mini (gpt-4o-mini-2024-07-18) to perform data cleaning.[30] We constructed a custom prompt that instructed the LLM to grade the quality of the ground truth reasonings based on whether it contained explanations for both correct and incorrect answers, explanations for only correct answers, or no relevant explanations at all. An example of this prompt is depicted in **Supplemental Figure 2**. Based on this workflow, we excluded 920 MCQ items which did not have any relevant explanations. Additionally, for the purpose of standardising evaluation items across the four models, we excluded 182 MCQ items where DeepSeek-R1 failed to complete processing due to timeouts or incomplete outputs. The final dataset consisted of 5,888 MCQ items, of which 2,729 items contained explanations for both correct and incorrect answers.

## Prompt Engineering and Hyperparameters

To minimize variability, we standardized the input prompts using a role-play format, instructing the LLMs to respond as an "assistant specializing in ophthalmology" (**Supplemental Figure 3**). All MCQ items, including questions and the corresponding options, were reformatted into structured prompts. Each prompt was then independently presented to each LLM through their respective APIs in a zero-shot setting. The models were then instructed to select the most appropriate answer, provide an explanation for the choice, and format their final response in JSON-compatible style.

To simulate real-world usage, we utilized the default hyperparameters for all LLMs.[31] The temperature, which is a parameter that regulates the randomness of the LLM's outputs, was set to 1 for OpenAI o1, o3-mini and DeepSeek-R1, and 0.7 for Gemini 2.0 Flash-Thinking.[32] In addition, for the OpenAI o1 and o3-mini, the "reasoning_effort" parameter was set to the default value of "medium".

## Evaluation Metrics

We evaluated the outputs of the four reasoning-focused LLMs using accuracy- and reasoning-based metrics. The accuracy of each LLM was evaluated by its overall accuracy, which was defined as the proportion of correct predictions made from all MCQ items. Additionally, the Macro-F1 score for each LLM was calculated as the unweighted harmonic mean of precision and recall over each MCQ option. We utilized bootstrapping with 1,000 repeats to determine the mean and 95% confidence interval of the Macro-F1 score for each LLM. In certain instances, the LLMs would select multiple or none of the options, which were treated as wrong predictions by default.

As the intermediate CoT of the OpenAI LLMs are intentionally concealed from end users,[33] we opted to not assess intermediate CoTs in a quantitative manner. Instead, we assessed the reasoning abilities of the LLMs based on the explanations they were prompted to provide for their choices. The reasoning abilities of each LLM were evaluated using five text-generation metrics: the Recall-Oriented Understudy for Gisting Evaluation (ROUGE-L), Metric for Evaluation of Translation with Explicit Ordering (METEOR), BERTScore, BARTScore, and AlignScore.[34-38] These metrics evaluate the quality of each LLM's final output by comparing it to the ground truth reasoning for each MCQ item; the intermediate thinking process of the models were not included in this analysis. In short, ROUGE-L measures the longest sequence of overlapping words, while METEOR evaluates the number of similar words between both sequences. BERTScore assesses the similarity between embeddings generated by the BERT model of both sequences, and BARTScore measures how effectively the BART model can predict each sequence from the other. Lastly, AlignScore measures the overall factual consistency of information between each sequence. A detailed explanation of these metrics is included in **Supplementary Table 1**. For each metric, we calculated the mean for each LLM and obtained the 95% confidence intervals for the means through bootstrapping with 1,000 repeats. Furthermore, we normalized each metric (scaling from 0 to 1) to facilitate comparison.

Additionally, to assess the inference time efficiency of each LLM, we randomly chose a subset of the MedMCQA dataset (n = 100) and recorded the running time required for each LLM to generate an output for each input prompt.

**Subgroup Evaluations**

We performed an additional evaluation on a subgroup consisting of only MCQ items with ground truth reasonings for both correct and incorrect answers (n = 2,729). This was performed to further assess the LLMs' ability to justify reasons for not selecting the incorrect answers. Further subgroup analyses by sub-specialties were also performed to identify any potential variation in the LLMs' performance across different subdomains in ophthalmology. Using the category labels provided by the MedMCQA dataset, we further stratified subgroups of QA items pertaining to cataracts (n = 330), retinal diseases (n = 307), and glaucoma (n = 208).

## Qualitative Evaluation

To complement the quantitative text-generation metrics, two board-certified ophthalmologists (MZ, GY) further performed qualitative evaluation on selected LLM thinking processes and outputs. A total of 300 MCQs involving differential diagnoses were randomly selected for review. Since OpenAI models do not provide their intermediate CoT reasoning, the summaries displayed in the ChatGPT interface were assessed instead. The evaluation focused on clarity, completeness, and structure of reasoning.

## Statistical Analysis

To compare accuracy at the item level, the two-tailed paired t-test was used to determine statistical significance between every pair of LLMs. For comparisons of bootstrapped Macro-F1 score, z-tests were applied, with Bonferroni correction used to control for multiple test comparisons.

For the text-generation metrics, the two-tailed Wilcoxon rank-sum test was used to determine statistically significant differences between the LLMs. This test was chosen due to the non-parametric nature of the text-generation metrics' output value distributions. Bonferroni correction was also applied to adjust for multiple comparisons.

A *p* value of <0·05 (post-Bonferroni correction) was deemed as statistically significant. All statistical analyses were conducted using Python (Python version 3.10.12, Python Software Foundation).

## Results

### Accuracy Evaluation

**Figure 3** and **Table 2** show the performance of the four reasoning-focused LLMs in terms of accuracy and Macro-F1 score. For accuracy, DeepSeek-R1 (0·888) and OpenAI o1 (0·902) both surpassed OpenAI o3-mini (0·857) and Gemini 2.0 Flash-Thinking (0·806) (all $p < 0.001$). However, there was no statistical difference between DeepSeek-R1 and OpenAI o1 ($p = 0.083$). In terms of Macro-F1 score, OpenAI o1 (0·900, 95% CI: 0·893 - 0·908) had the highest score, followed by DeepSeek-R1 (0·885, 95% CI: 0·877 - 0·893), OpenAI o3-mini (0·855, 95% CI: 0·847 - 0·865) and Gemini 2.0 Flash-Thinking (0·804, 95% CI: 0·793 - 0·814) (all $p < 0.001$).

**Reasoning Evaluation**

**Figures 4-5** and **Table 2** highlight the relative strengths and weaknesses of all four LLMs in reasoning capabilities, illustrating inconsistencies in the four LLMs' performance across different text-generation metrics. In ROUGE-L score, OpenAI o3-mini (0·151, 95% CI: 0·150 - 0·153) outperformed others (all $p < 0.001$). Based on the METEOR score, OpenAI o1 (0·232, 95% CI: 0·229 - 0·234) excelled in comparison to the other models (all $p < 0.001$). For BERTScore, DeepSeek-R1 (0·673, 95% CI: 0·672 - 0·675) and OpenAI o3-mini (0·673, 95% CI: 0·672 - 0·674) were tied and both were better than o1 and Gemini 2.0 Flash-Thinking (all $p < 0.001$). For BARTScore, DeepSeek-R1 (-4·105, 95% CI: -4·128 - -4·080) and Gemini 2.0 Flash-Thinking (-4·126, 95% CI: -4·149 - -4·104) emerged as joint-best (all $p < 0.001$ versus OpenAI o1 and OpenAI o3-mini). Finally, based on AlignScore, o3-mini attained the highest score (0·181, 95% CI: 0·177 - 0·185), significantly outperforming DeepSeek-R1 and Gemini 2.0 Flash-Thinking (both $p < 0.001$).

**Inference Time Analysis**

The average time needed for each reasoning-focused LLM to generate a response to an input prompt is presented in **Figure 6.** DeepSeek-R1 required the longest processing time (40·3 seconds on average per output, $p < 0.001$ versus all others), followed by OpenAI o1 (17·3s, $p < 0.001$ vs o3-mini and Gemini 2.0 Flash Thinking). OpenAI o3-mini (8·8s) and Gemini 2.0 Flash-Thinking (6·7s) took the least amount of time to generate outputs.

**Subgroup Evaluation on Items with Reasonings for Correct and Incorrect Answers**

For the subgroup consisting of MCQ items with gold-standard reasonings for both correct and incorrect answers (n = 2,729), the performance trends mirrored those observed in the full dataset. As shown in **Table 3** and **Supplemental Figure 4**, OpenAI o1 (0·901) remained the best in accuracy, followed by DeepSeek-R1 (0·886), with no significant between their performances ($p$ = 0·477). Similarly, OpenAI o1 achieved the best performance in Macro-F1 (0·900, 95% CI: 0·889 - 0·912), outperforming all other models (all $p$ < 0·001), followed by DeepSeek-R1 (0·885, 95% CI: 0·872 - 0·896), o3-mini (0·855, 95% CI: 0·842 - 0·869) and Gemini 2.0 Flash-Thinking (0·788, 95% CI: 0·773 - 0·802).

In terms of the text-generation metrics (**Table 3**, **Supplemental Figure 5** and **Supplementary Figure 6**), OpenAI o3-mini outperformed all other LLMs for ROUGE-L (0·172, 95% CI: 0·170 - 0·174, $p$ < 0·001 versus all others). Based on METEOR, OpenAI o1 outperformed all other LLMs (0·247, 95% CI: 0·244 - 0·250, $p$ < 0·001 for all). DeepSeek-R1 performed the best in BERTScore (0·689, 95% CI: 0·688 - 0·691, $p$ < 0·001 versus all others), and was joint-best with Gemini 2.0 Flash-thinking in terms of BARTScore (DeepSeek-R1: -4·091, 95% CI: -4·118 - -4·063, Gemini 2.0 Flash-Thinking: -4·107, 95% CI: -4·134 - -4·079, $p$ = 2·237). Finally, OpenAI o3-mini (0·244, 95% CI: 0·239 - 0·249) and OpenAI o1 (0·232, 95% CI: 0·227 - 0·237) were joint-best in AlignScore.

**Subgroup Analyses Based on Ophthalmology Sub-specialties**

Performance trends across three ophthalmology sub-specialties (cataract, retina, and glaucoma) were consistent with the main analysis (**Supplemental Tables 2-4, Supplementary Figures 7-9**). OpenAI o1 surpassed all other LLMs in terms of accuracy and Macro-F1 in all three subspecialties. When it comes to text-generation metrics, o3-mini consistently performed the best in ROUGE-L and AlignScore, while o1 also excelled in METEOR. For BERTScore, DeepSeek-R1 performed the best in all subtopics except for "Glaucoma", where o3-mini performed the best. DeepSeek-R1 also performed the best for

BARTScore in all subtopics except for the "Glaucoma" subtopic, where Gemini 2.0 Flash-Thinking performed the best.

**Qualitative Assessment of LLMs' Thinking Process and Final Outputs**

Examples of the responses (both the CoT thinking process and final output) by the four reasoning-focused LLMs are shown in **Supplemental Tables 5-7**. Regarding the intermediate thinking processes generated by the four reasoning-focused LLMs, we observed that the internal reasoning of DeepSeek-R1 exhibited a more human-like reasoning approach. Its outputs often began with an initial diagnostic impression and revealed moments of self-questioning and uncertainty, with phrases such as "wait, another thought" and "but I am a bit confused here". In contrast, the OpenAI models displayed more succinct and streamlined reasoning processes with less 'self-deliberation'. On the other hand, Gemini 2.0 Flash-Thinking provided systematic, methodically structured reasoning, offering a more logical progression compared to DeepSeek-R1. Furthermore, Gemini's thinking process and outputs share a similar structured format, with both organized by options.

In terms of final outputs, DeepSeek-R1 tended to present the correct answer at the very start of its response, providing clarity and ease of comprehension for users. However, because the reasoning had already been embedded in its transparent thinking process, the depth of explanations in its outputs was often relatively limited (**Supplemental Tables 5-7**). On the other hand, OpenAI's o1 and Gemini 2.0 Flash-Thinking both delivered well-structured responses, providing comprehensive outputs. However, Gemini 2.0 Flash-Thinking tended to appear verbose and repetitive, even for simpler cases. O3-mini, meanwhile, focused primarily on identifying the correct answer - it often offered detailed reasoning for the correct options but was less consistent in explaining incorrect options, sometimes offering only general justifications.

## Discussion

Using a large dataset of 5,888 MCQ items with accompanying comprehensive ground truth reasonings, we performed a thorough evaluation of four reasoning-focused LLMs – DeepSeek-R1, OpenAI o1, OpenAI o3-mini, and Gemini 2.0 Flash-Thinking. While OpenAI o1 and DeepSeek-R1 demonstrated superior accuracy, the LLMs' performances across text-generation metrics varied, with each model excelling in different aspects of reasoning quality. By integrating large-scale benchmarking with both quantitative performance metrics and qualitative analysis, our study presents a multi-dimensional evaluation framework for assessing reasoning-focused LLMs in ophthalmology. To our knowledge, this is the first study to apply such a comprehensive and systematic evaluation approach to these relatively new reasoning-focused LLMs in ophthalmology. Unlike recent reasoning LLM benchmarking studies that primarily focus on single model assessments (mainly DeepSeek-R1) and rank models solely based on accuracy,[39,40] our study adopts a broader evaluation strategy. We provided a nuanced analysis of each model's strengths, limitations, and trade-offs, offering new insights into the evolving role of reasoning-focused LLMs in medical AI.

Despite the strong performances of all evaluated LLMs in terms of accuracy (with o1 and DeepSeek-R1 being comparatively superior), our quantitative reasoning assessment revealed inconsistencies across five text-generation metrics, each capturing distinct dimensions of reasoning quality relative to the MedMCQA ground truths. Notably, no single model emerged as universally superior across all metrics, underscoring the complexity of evaluating LLM reasoning. O3-mini ranked the highest in ROUGE-L, which compares the exact word order in the longest common sequence between the ground truth and the model output (**Figure 4 and Supplementary Table 1**). This indicates that o3-mini's responses most closely mirrored the structure and phrasing of the ground truth reasonings in MedMCQA. Meanwhile, o1 led in METEOR, which considers synonym recognition and word-form variants, suggesting that it maintains meaning even when lexical choices differ. Both Deepseek-R1 and o3-mini excelled in BERTScore, demonstrating strong semantic similarity to reference answers. However, in BARTScore where fluency is considered along with semantic similarity, o1 outperformed the others, indicating that its responses are meaningfully accurate and naturally phrased. Finally, o3-mini ranked highest in AlignScore, suggesting that its outputs have the strongest factual consistency with the key ideas in the ground truth reasonings. These findings highlight the multi-faceted nature of reasoning evaluation in LLMs, demonstrating that while o3-mini and o1 perform well across multiple metrics, no single benchmark fully captures the complexity of reasoning quality. Thus, a comprehensive, multi-metric evaluation framework remains essential for accurately assessing LLM reasoning performance in specialized medical domains.

One caveat of using text-generation metrics is that they are ground truth dependent and hence may lack sufficient context awareness to fully capture reasoning performance. Therefore, to gain deeper insights into each LLM's thinking process and output response, we further conducted a qualitative evaluation on questions requiring differential diagnosis, a task that demands high-level reasoning. The analysis focused on identifying similarities and differences across the four LLMs. Both OpenAI o1 and Gemini 2.0 Flash-Thinking provided in-depth and structured responses, which could be attributed to the enhanced "chain-of-thought" reasoning architecture of o1 and step-by-step reasoning approach of Gemini 2.0 Flash-Thinking.[11,18] For instance, for a question that involved the differential diagnosis of normal tension glaucoma (NTG, **Supplemental Table 7**), both o1 and Gemini 2.0 Flash-Thinking started by identifying the key features of the question that matched characteristics of NTG (normal intraocular pressure, optic disc cupping and defected visual field). They then gave a detailed explanation of the incorrect options, followed by their rationales for eliminating these options from consideration. This level of thoroughness was notably beneficial for more complex topics. The thinking process of DeepSeek-R1 were very detailed, sometimes re-visiting previously-made points, which had the effect of making it seem quite long-winded. However, its responses were less in-depth than the previous LLMs, often without a detailed rationale for the incorrect options. Finally, o3-mini tended to present straightforward and concise responses, providing reasons for both correct and incorrect options in an efficient manner. As such, o3-mini may be suitable for tasks or scenarios that prioritize efficiency and direct, high-yield responses over exhaustive explanations.

The difference in reasoning patterns of reasoning-focused LLMs could be due to their unique characteristics **(Figure 2)**: OpenAI models do not display their full intermediate reasoning process, therefore making their displayed thinking processes appear more concise but less transparent.[41] On the other hand, both DeepSeek-R1 and Gemini 2.0 Flash-Thinking show each step of their thorough and detailed thinking processes, thus providing greater explainability and transparency. Due to DeepSeek-R1's imitation of human-like thinking processes, its approach to elimination involves numerous instances of reconsideration and self-questioning, thus presenting as less organized compared to the other three models.[18,28] Ultimately, each reasoning-focused LLM model caters to different task or clinical operational needs. Some scenarios demand a concise reference that can be quickly scanned for immediate clarity, whereas others call for a more thorough explanation that addresses nuances, context, and complexity. In contexts where immediate responses to simple queries are necessary, the multi-step reasoning abilities of these LLMs may introduce unnecessary delay, making non-reasoning LLMs a more practical choice for these instances. [42]

When considering potential clinical deployment, we observed certain strengths and weaknesses for each reasoning-focused LLM. DeepSeek-R1 stands out as being open-weight, making it the only reasoning-type LLM that can be locally hosted and customized by healthcare institutes.[43] However, the time required for DeepSeek-R1 to generate outputs was up to eight times longer than other LLMs' (**Figure 6**). Another limitation of DeepSeek-R1 was its inability to process some MCQ items, frequently timing out before generating a response. This resulted in significant time delays and inefficiencies. Upon further reviewing the 182 MCQs where DeepSeek-R1 encountered errors, no discernible pattern among the problematic questions was found, which may indicate that the timeouts might arise from server-related constraints. The longer processing time of DeepSeek-R1 may hamper its ability to be used practically in real-world time-sensitive clinical scenarios.[44,45] Meanwhile, both OpenAI LLMs had significantly higher financial costs than DeepSeek-R1, with o3-mini requiring twice the costs of DeepSeek-R1 (based on token costs for uncached inputs and outputs), and o1 requiring nearly fifteen times that of o3-mini.[28,41] Additionally, OpenAI hides the full intermediate thinking process from its models' outputs, instead displaying a model-generated summary of the CoT process. However, all intermediate tokens generated are still charged as full output tokens, which adds significant hidden deployment costs for the OpenAI LLMs. Finally, as of this article's submission, Gemini 2.0 Flash-Thinking is currently in an "experimental" stage, with undisclosed pricing and potential performance changes. [18,44,45]

The strengths of this study include the use of a large-scale dataset consisting of 5,888 ophthalmology-specific QA-items with corresponding ground truth reasonings. To further enhance the reliability of our dataset, we performed additional quality checking and filtering of the MCQ items, ensuring the validity of our text-generation metric-based evaluations. Additionally, to complement the text-generation metrics, we also assigned two board-certified ophthalmologists to independently review model-generated responses for differential diagnosis items, providing unique and clinically relevant insights into each LLM's reasoning style and capabilities. Nevertheless, this study also has several limitations. First, as OpenAI restricts access to the full intermediate thinking steps of its o1 and o3-mini models, we were only able to examine the model-generated reasoning summaries, rather than their complete internal thinking processes.[46] In addition, while the MedMCQA ophthalmological dataset serves as a valuable benchmark, it does not encompass all potential real-world applications and tasks in ophthalmological practice. Thus, future study on the real-world evaluation of these reasoning-focused LLMs is warranted.

The validity of the text-generation metrics used is dependent on the quality of the ground truth reasonings, underscoring the need for more robust ophthalmological-specific benchmarks that include high-quality ground truth reasonings. Moving forward, future research should focus on robust, quantitative human evaluations of the LLM's reasonings across varied clinical scenarios, so as to holistically validate their practical utility.

**Conclusion**

Reasoning-focused LLMs show early promise for ophthalmology. Nevertheless, across all evaluation dimensions, no single model consistently outperforms others. While o1 and DeepSeek-R1 achieved the highest accuracy, DeepSeek-R1's slow inference speed may limit its practicality in real-world applications. The lack of a clear best-performing model underscores the importance of multi-metric benchmarking and qualitative evaluation when assessing LLM's suitability for clinical deployment. Our findings collectively underscore the trade-offs between accuracy, reasoning quality, and processing speed, each of which must be carefully balanced when considering deployment of these next-generation reasoning-focused LLMs in clinical workflows.

**Figures**

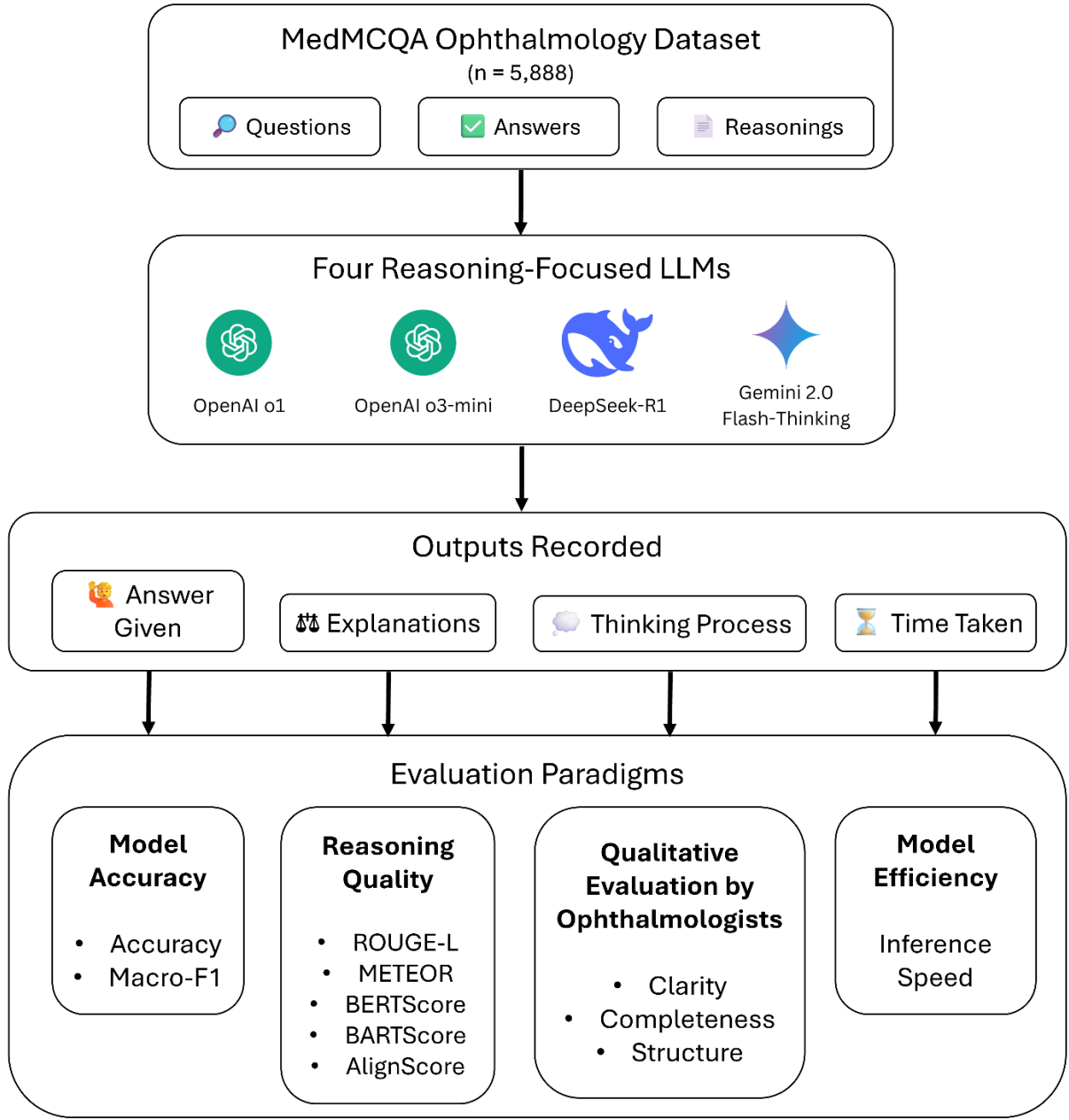

**Figure 1.** Overview of Study Design.

**Figure 2.** Examples of the User Interfaces and Reasoning Processes for each of the Four Reasoning-Focused LLMs

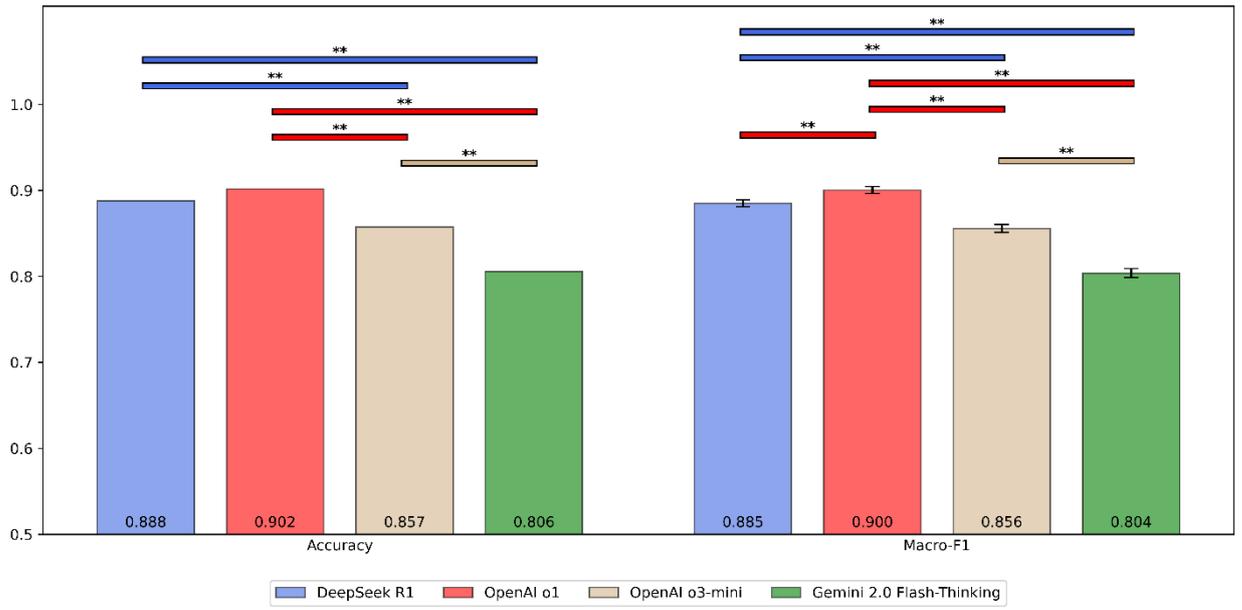

**Figure 3**. Performance of Accuracy and Macro-F1 for the Four Reasoning-Focused LLMs

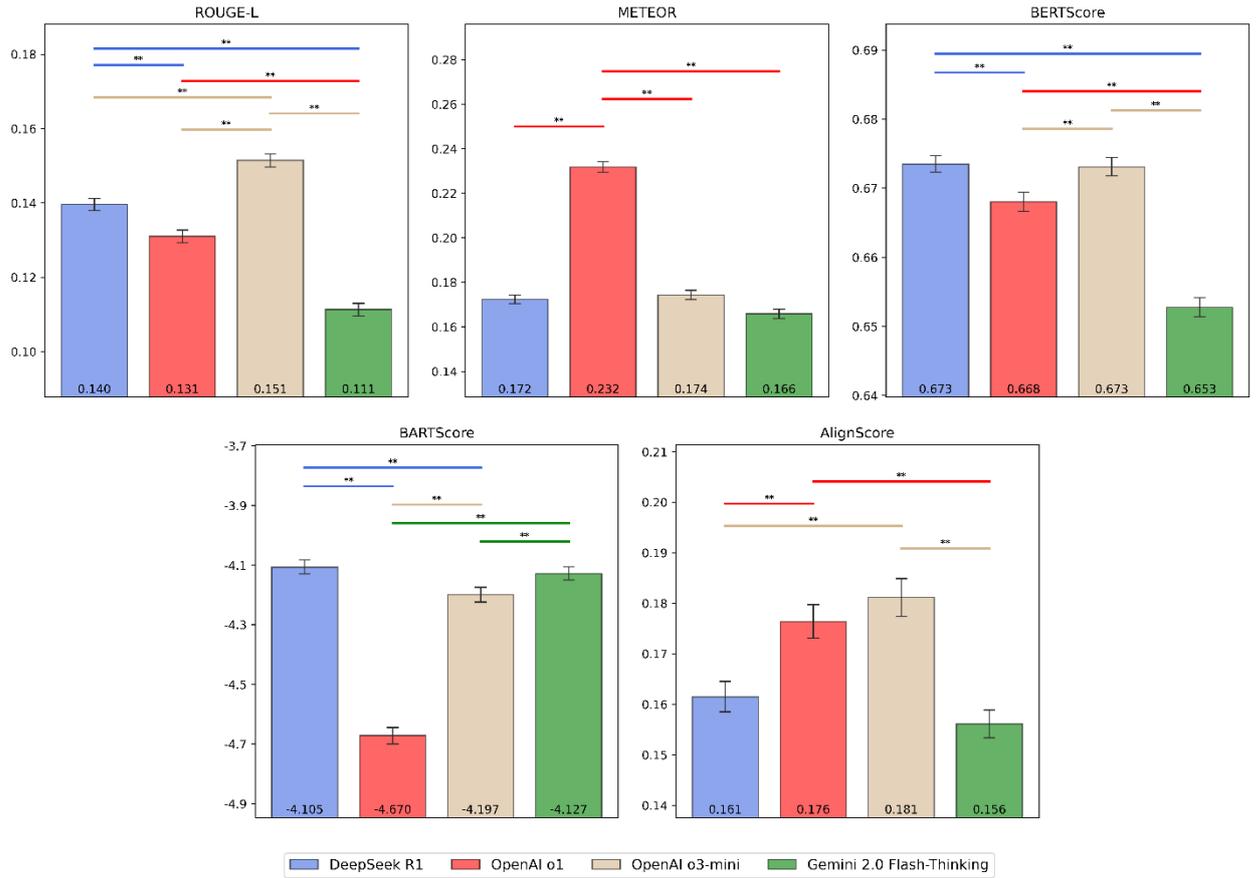

**Figure 4.** Reasoning Capabilities through Five Text-Generation Metrics for the Four Reasoning-Focused LLMs

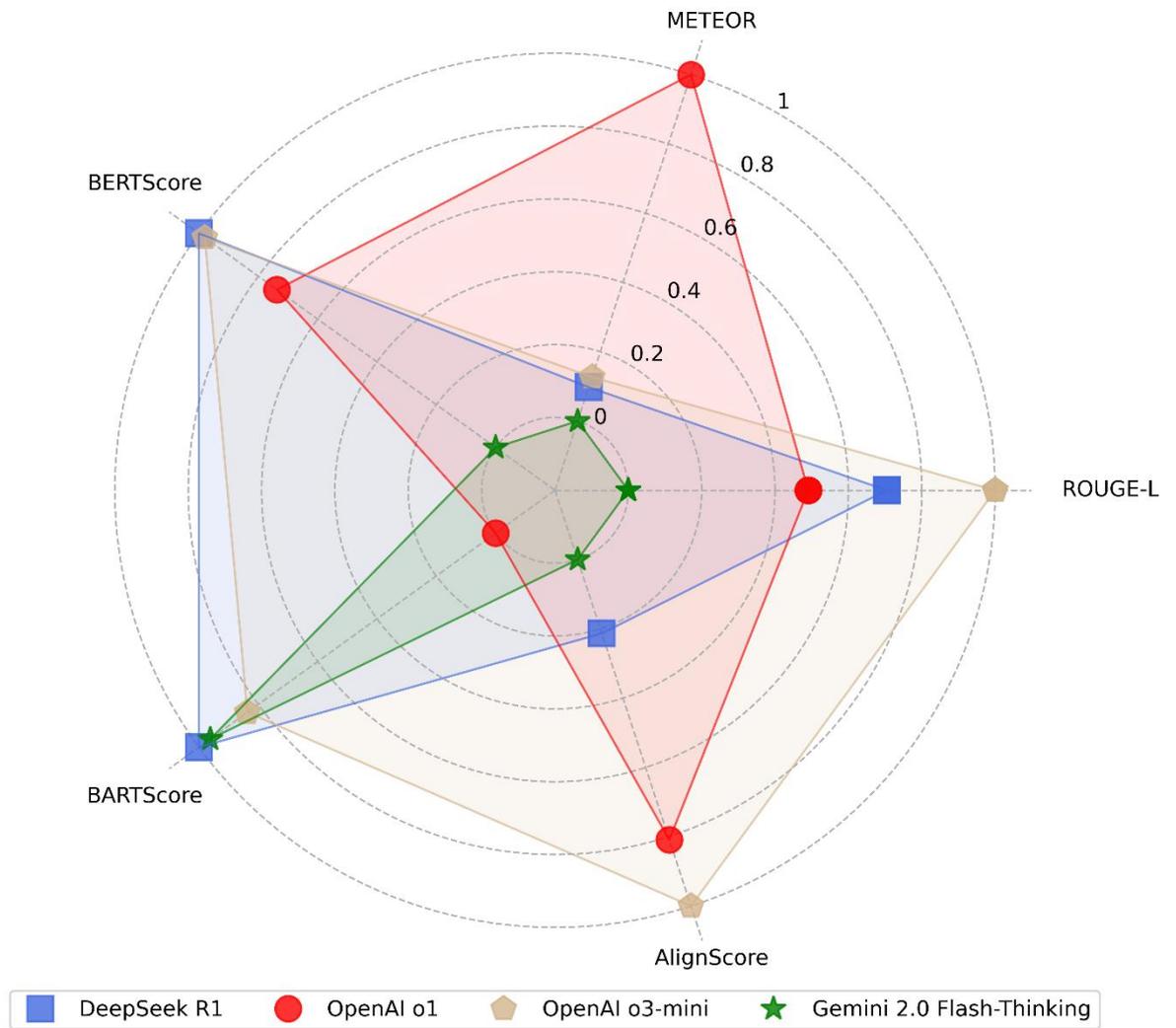

**Figure 5.** Radar Chart of Normalized Text-Generation Metrics for the Four Reasoning-Focused LLMs

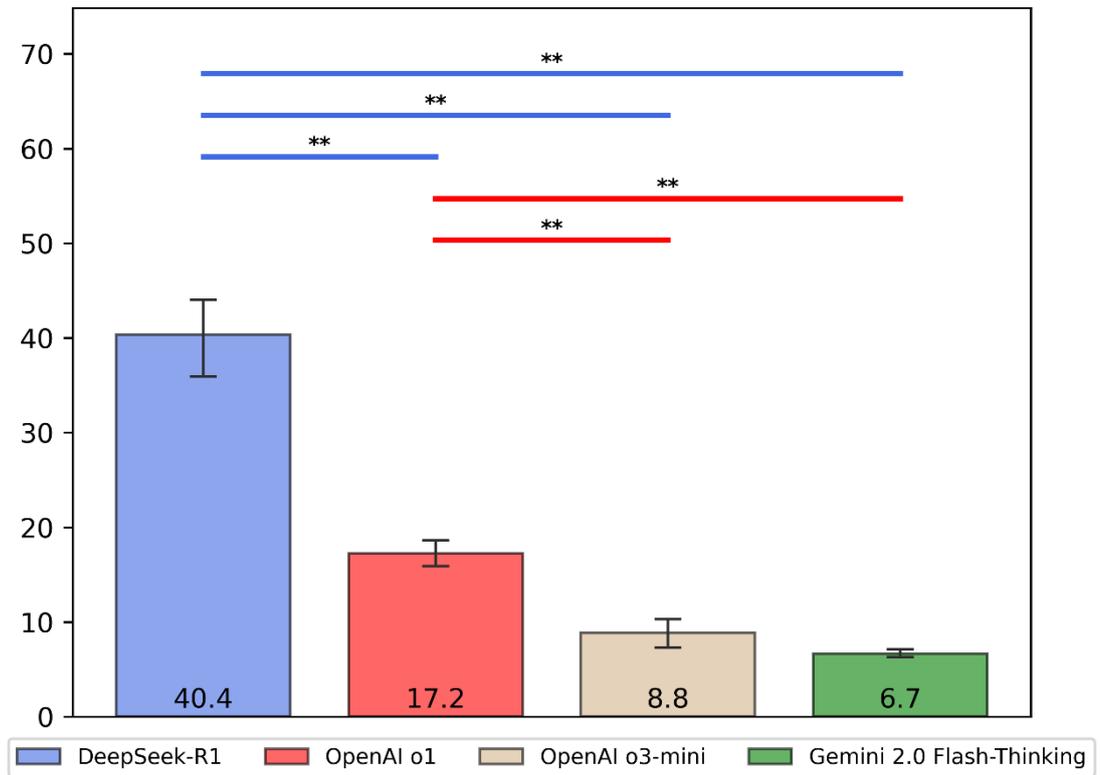

**Figure 6.** Inference Time (per item) Required for the Four Reasoning-Focused LLMs

**Tables**

**Table 1: Characteristics of Reasoning-Focused LLMs**

| | Reasoning-focused LLMs: | | | |
| --- | --- | --- | --- | --- |
| | **DeepSeek-R1** | **o1** | **o3-mini** | **Gemini 2.0 Flash- Thinking** |
| **Developer** | DeepSeek | OpenAI | OpenAI | Google |
| **Release Date** | Jan-25 | Sep-24 | Jan-25 | Dec-24 |
| **Architecture** | Mixture-of-Experts | Not specified | Not specified | Not specified |
| **Model Size/Context Window** | 671B, 37B active parameters per token/128K tokens | Not specified /200K tokens | Estimated to be around 200B/200K tokens | Not specified/1M tokens |
| **Reasoning Approach** | Visible reasoning steps ("thinking out loud") | Enhanced reasoning with "chain-of-thought" | Simulated reasoning with adjustable effort levels | Step-by-step reasoning through a dropdown menu |

|  |  | processing |  |  |
|---|---|---|---|---|

**Table 2**. Comparative Performance of DeepSeek-R1, OpenAI o1, o3-mini, and Gemini 2.0 Flash-Thinking on the full MedMCQA ophthalmology dataset (n = 5,888).

|  | Reasoning-focused LLMs: | | | |
|---|---|---|---|---|
| **Metrics:** | **DeepSeek R1** | **OpenAI o1** | **OpenAI o3-mini** | **Gemini 2.0 Flash-Thinking** |
| **Accuracy** | 0·888 | **0·902** | 0·857 | 0·806 |
| **Macro-F1** | 0·885 (0·877 - 0·893) | **0·900 (0·893 - 0·908)** | 0·856 (0·847 - 0·865) | 0·804 (0·793 - 0·814) |
| **ROUGE-L** | 0·140 (0·138 - 0·141) | 0·131 (0·129 - 0·133) | **0·151 (0·150 - 0·153)** | 0·111 (0·110 - 0·113) |
| **METEOR** | 0·172 (0·170 - 0·174) | **0·232 (0·229 - 0·234)** | 0·174 (0·172 - 0·176) | 0·166 (0·164 - 0·168) |
| **BERTScore** | **0·673 (0·672 - 0·675)** | 0·668 (0·667 - 0·669) | 0·673 (0·672 - 0·674) | 0·653 (0·651 - 0·654) |
| **BARTScore** | **-4·105 (-4·128- -4·080)** | -4·670 (-4·698- -4·643) | -4·197 (-4·223- -4·173) | -4·127 (-4·149- -4·104) |
| **AlignScore** | 0·162 (0·158 - 0·165) | 0·176 (0·173 - 0·180) | **0·181 (0·177 - 0·185)** | 0·156 (0·153 - 0·159) |

**Table 3. Comparative Performance of DeepSeek-R1, OpenAI o1, o3-mini, and Gemini 2.0 Flash-Thinking on the MedMCQA ophthalmology subset with gold-standard reasonings for both correct and incorrect answers (n = 2,729).**

|  | Reasoning-focused LLMs: | | | |
|---|---|---|---|---|
| **Metrics:** | **DeepSeek R1** | **OpenAI o1** | **OpenAI o3-mini** | **Gemini 2.0 Flash-Thinking** |
| **Accuracy** | 0·886 | **0·901** | 0·856 | 0·787 |
| **Macro-F1** | 0·885 (0·872 - 0·896) | **0·900 (0·889 - 0·912)** | 0·856 (0·842 - 0·869) | 0·788 (0·773 - 0·802) |
| **ROUGE-L** | 0·164 (0·162 - 0·166) | 0·158 (0·156 - 0·160) | **0·172 (0·170 - 0·174)** | 0·140 (0·138 - 0·142) |
| **METEOR** | 0·174 (0·172 - 0·177) | **0·247 (0·244 - 0·250)** | 0·171 (0·168 - 0·174) | 0·191 (0·189 - 0·194) |
| **BERTScore** | **0·689 (0·688 - 0·691)** | 0·685 (0·684 - 0·687) | 0·686 (0·685 - 0·688) | 0·674 (0·673 - 0·676) |
| **BARTScore** | **-4·091 (-4·118 - -4·063)** | -4·697 (-4·726 - -4·669) | -4·180 (-4·205 - -4·155) | -4·107 (-4·134 - -4·079) |
| **AlignScore** | 0·215 (0·210 - 0·220) | 0·232 (0·227 - 0·237) | **0·244 (0·239 - 0·249)** | 0·209 (0·205 - 0·213) |

**Supplementary Materials**

Supplemental Figure 1. Flowchart for Dataset Curation Process

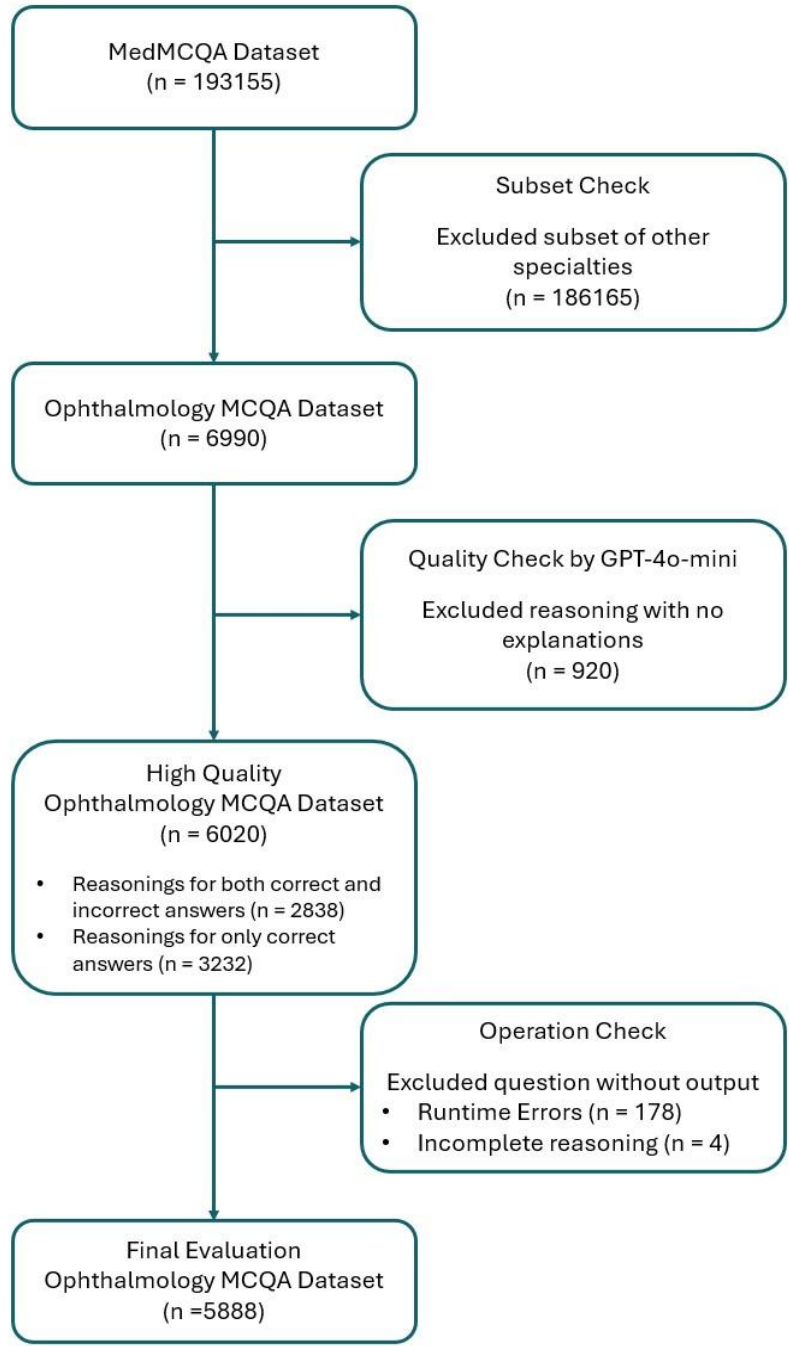

Supplemental Figure 2. Example of Prompt Used for GPT-4o-Mini-Based Evaluation of Ground Truth Reasoning Quality

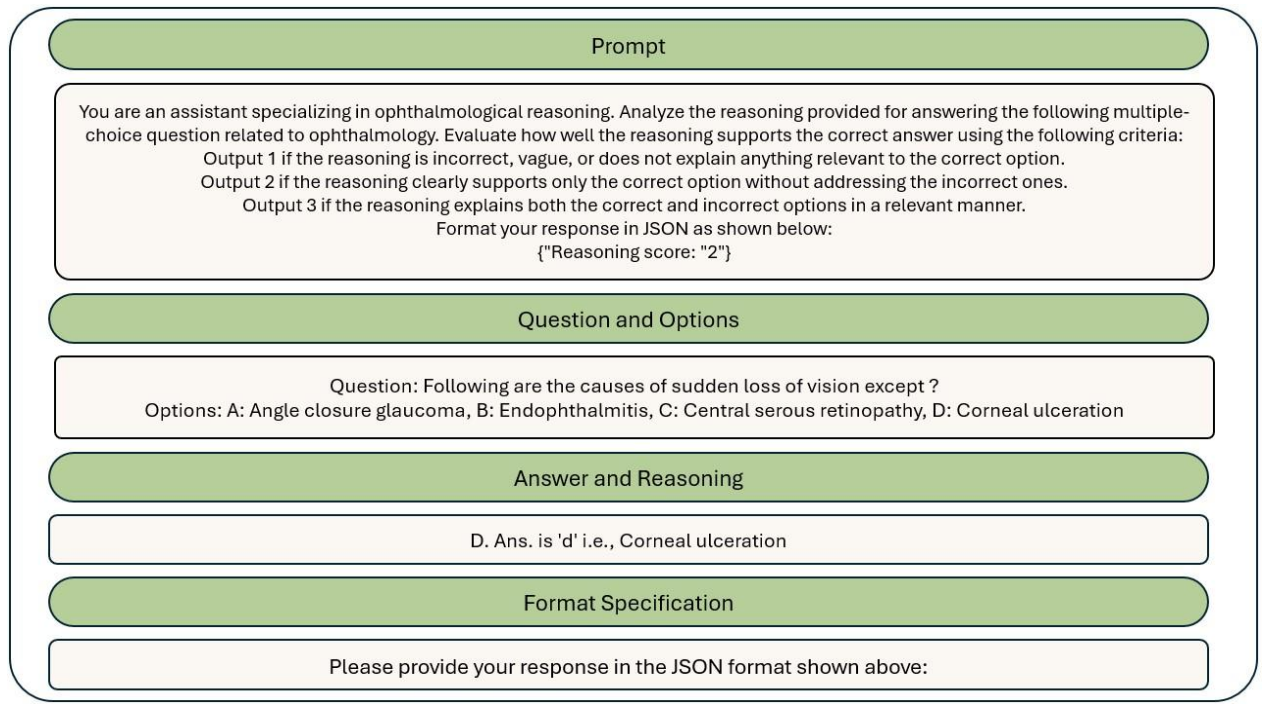

Supplemental Figure 3. Example of Prompt Used for Zero-Shot LLM Question-Answering and Reasoning

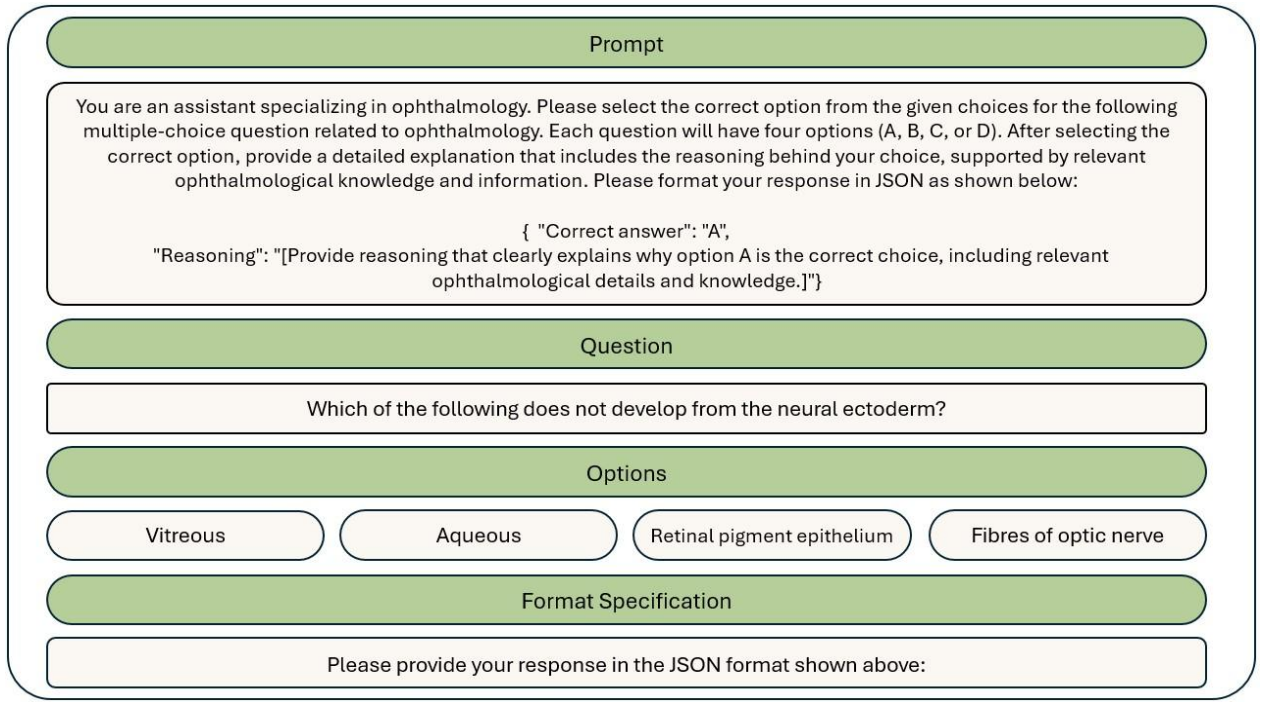

Supplemental Figure 4. Performance of Accuracy and Macro-F1 for the Four Reasoning-Focused LLMs on the MedMCQA Subset with Ground Truth Reasonings for Correct and Incorrect Answers (n = 2,729)

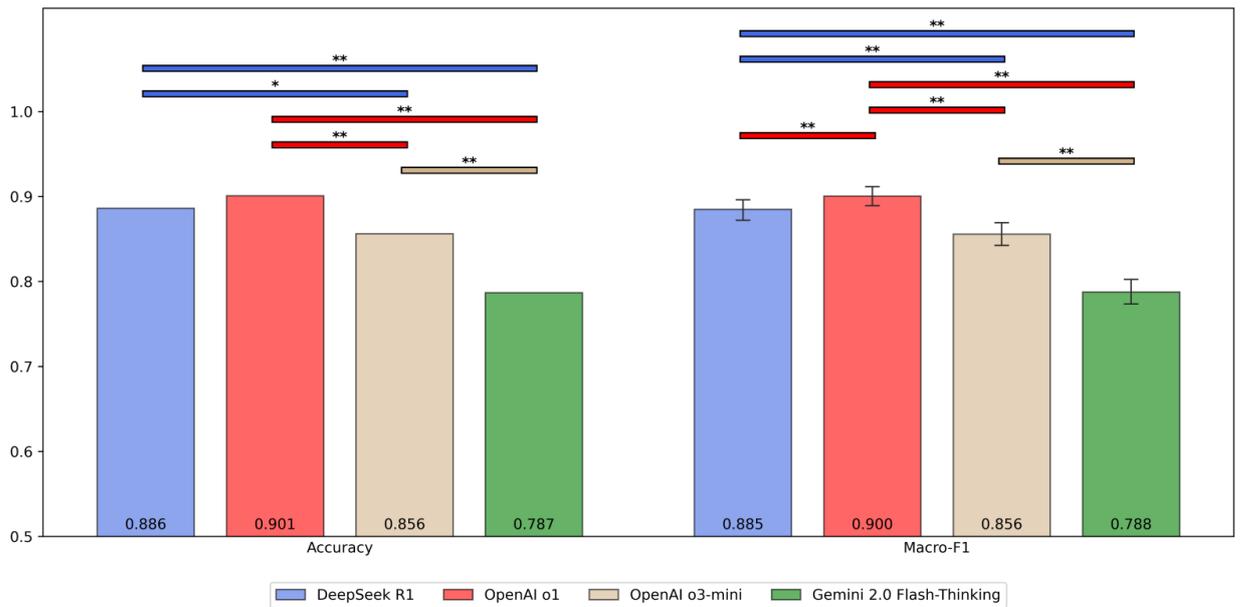

Horizontal bars indicate statistically significant differences between pairs of LLMs, with the color of the bar representing the LLM that performed better. Error bars for macro-F1 indicate 95% confidence intervals obtained from bootstrapping. * denotes p < 0·05, ** denotes p < 0·001, both after Bonferroni correction.

Supplemental Figure 5. Performance in Reasoning Capabilities through Five Text-Generation Metrics for the Four Reasoning-Focused LLMs on the MedMCQA Subset with Ground Truth Reasonings for Correct and Incorrect Answers (n = 2,729)

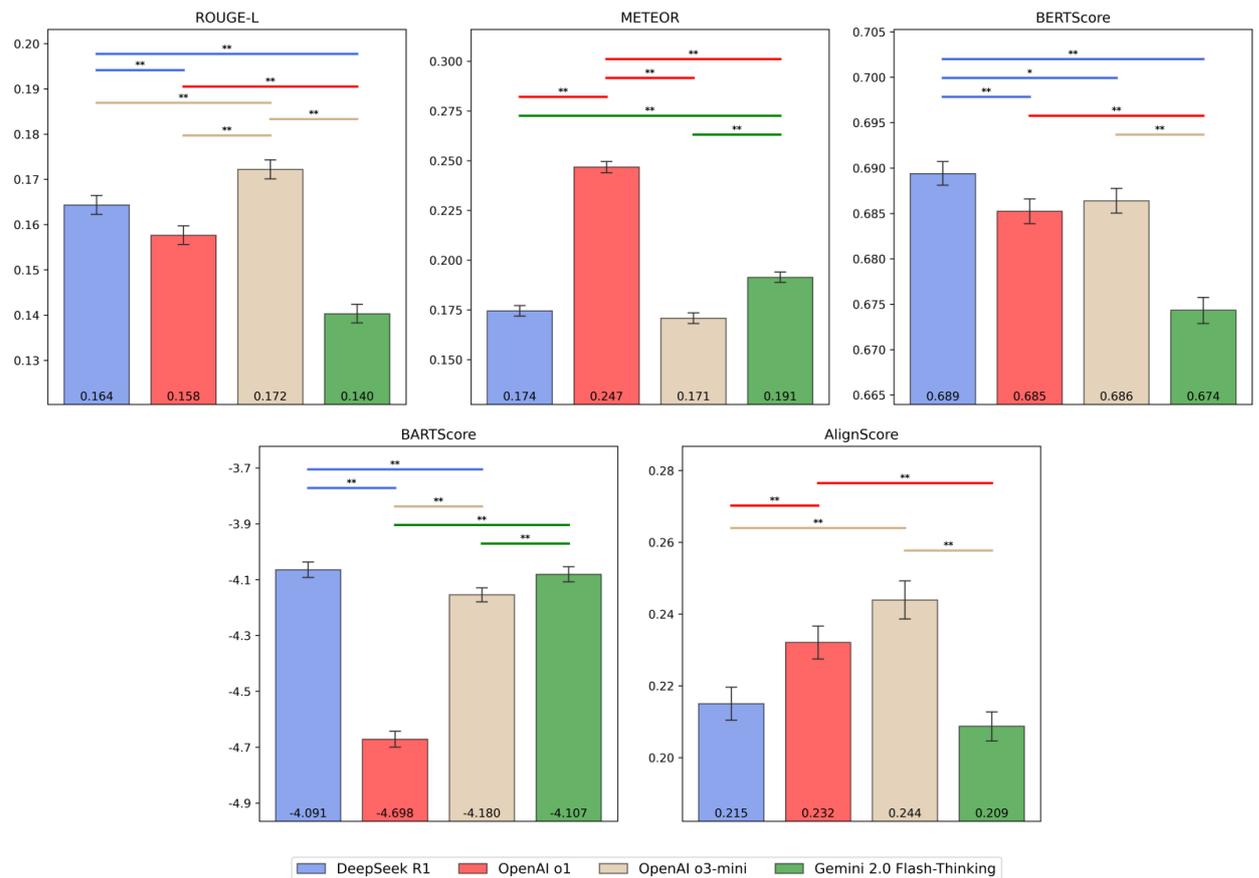

Horizontal bars indicate statistically significant differences between pairs of LLMs, with the color of the bar representing the LLM that performed better. Error bars for each metric indicate 95% confidence intervals obtained from bootstrapping. * denotes p < 0·05, ** denotes p < 0·001, both after Bonferroni correction.

Supplemental Figure 6. Radar Chart of Normalized Text-Generation Metrics for the Four Reasoning-Focused LLMs on the MedMCQA Subset with Ground Truth Reasonings for Correct and Incorrect Answers (n = 2,729)

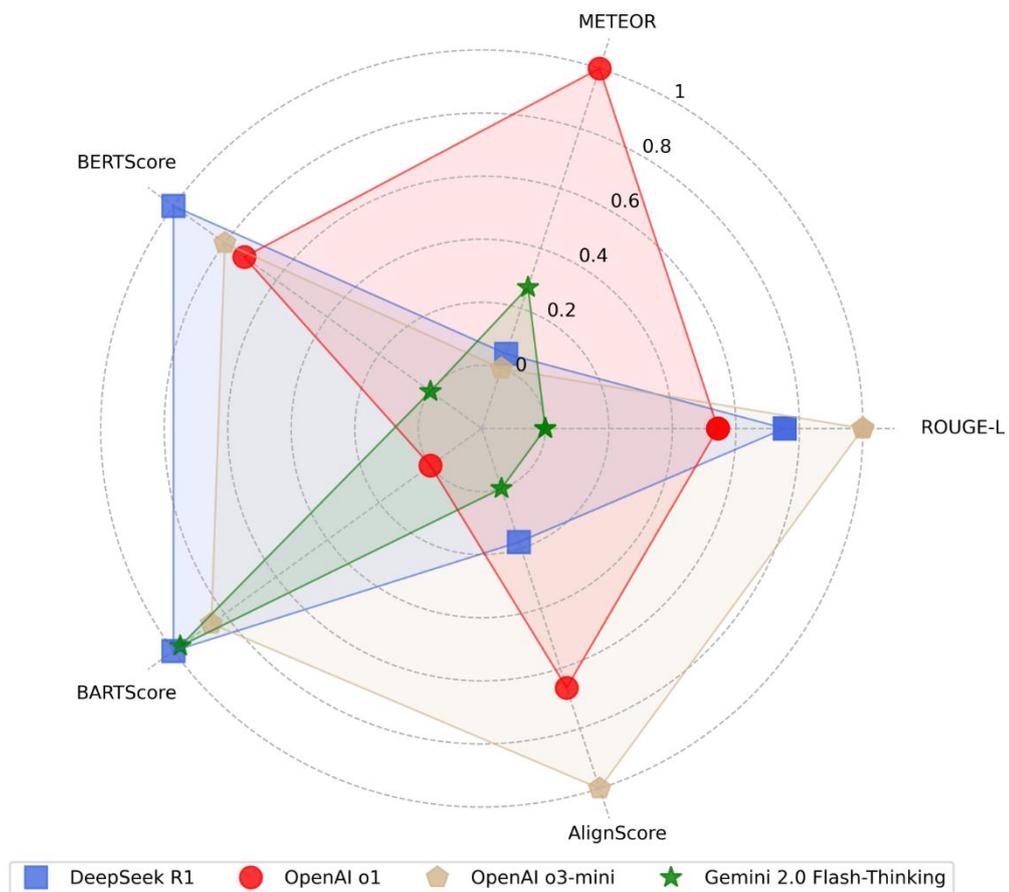

LLM performances for each metric were normalized to a scale from 0 to 1, with 1 representing the best-performing LLM, and 0 representing the worst-performing LLM. All other LLMs' performances were scaled linearly in-between.

Supplemental Figure 7. Reasoning Capabilities through Five Text-Generation Metrics for the Four Reasoning-Focused LLMs on the MedMCQA Subset with Question Items Relating to "Cataracts" (n = 330)

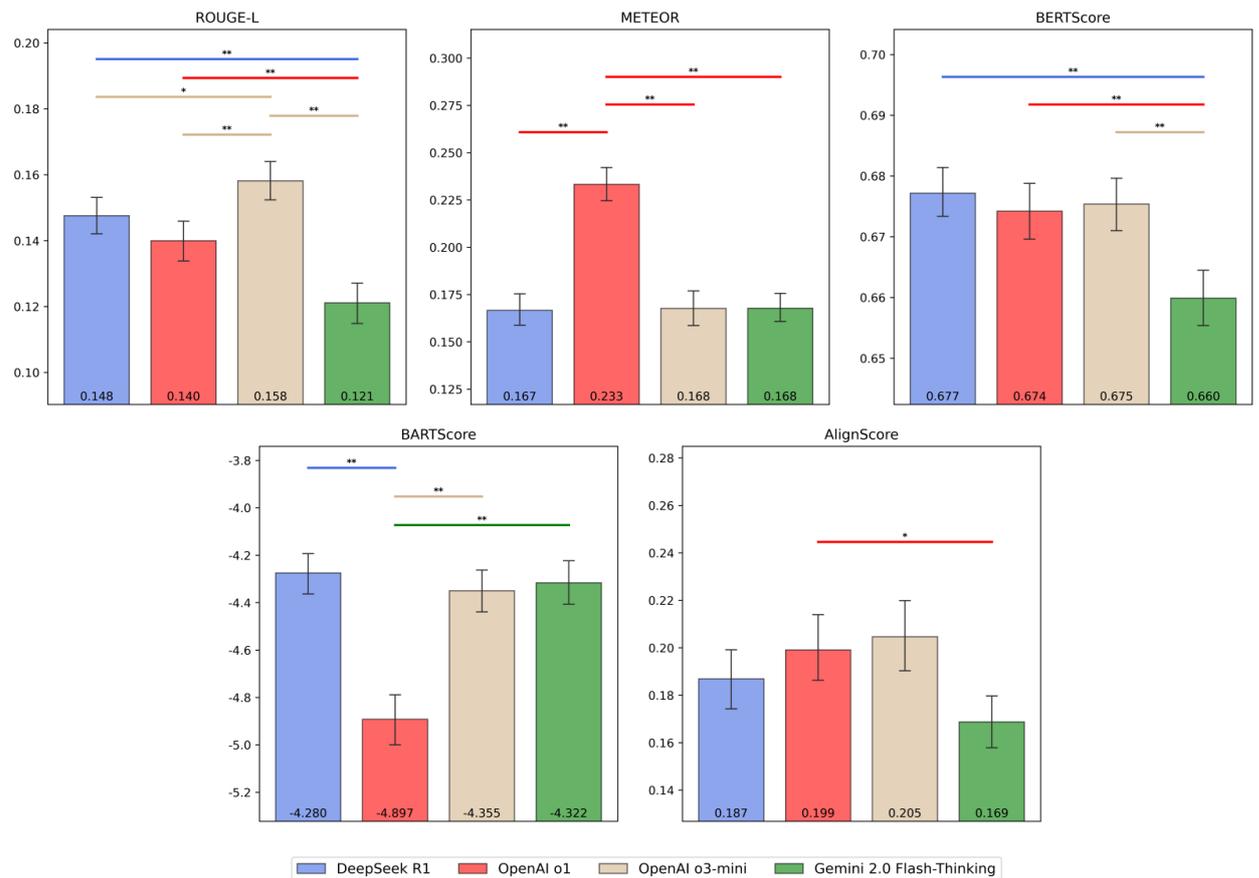

Horizontal bars indicate statistically significant differences between pairs of LLMs, with the color of the bar representing the LLM that performed better. Error bars for each metric indicate 95% confidence intervals obtained from bootstrapping. * denotes p < 0·05, ** denotes p < 0·001, both after Bonferroni correction.

Supplemental Figure 8. Reasoning Capabilities through Five Text-Generation Metrics for the Four Reasoning-Focused LLMs on the MedMCQA Subset with Question Items Relating to "Retinal Diseases" (n = 307)

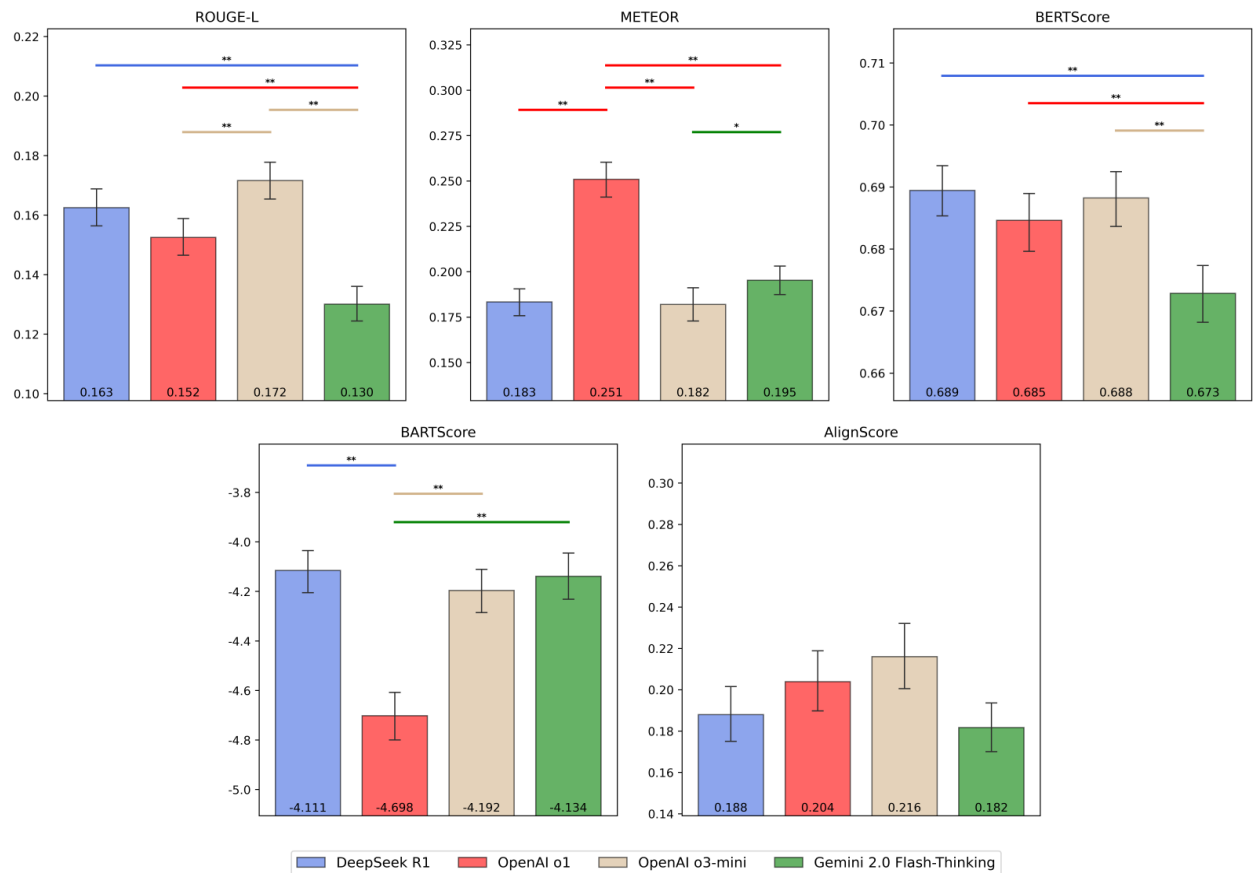

Horizontal bars indicate statistically significant differences between pairs of LLMs, with the color of the bar representing the LLM that performed better. Error bars for each metric indicate 95% confidence intervals obtained from bootstrapping. * denotes $p < 0.05$, ** denotes $p < 0.001$, both after Bonferroni correction.

Supplemental Figure 9. Reasoning Capabilities through Five Text-Generation Metrics for the Four Reasoning-Focused LLMs on the MedMCQA Subset with Question Items Relating to "Glaucoma" (n = 208)

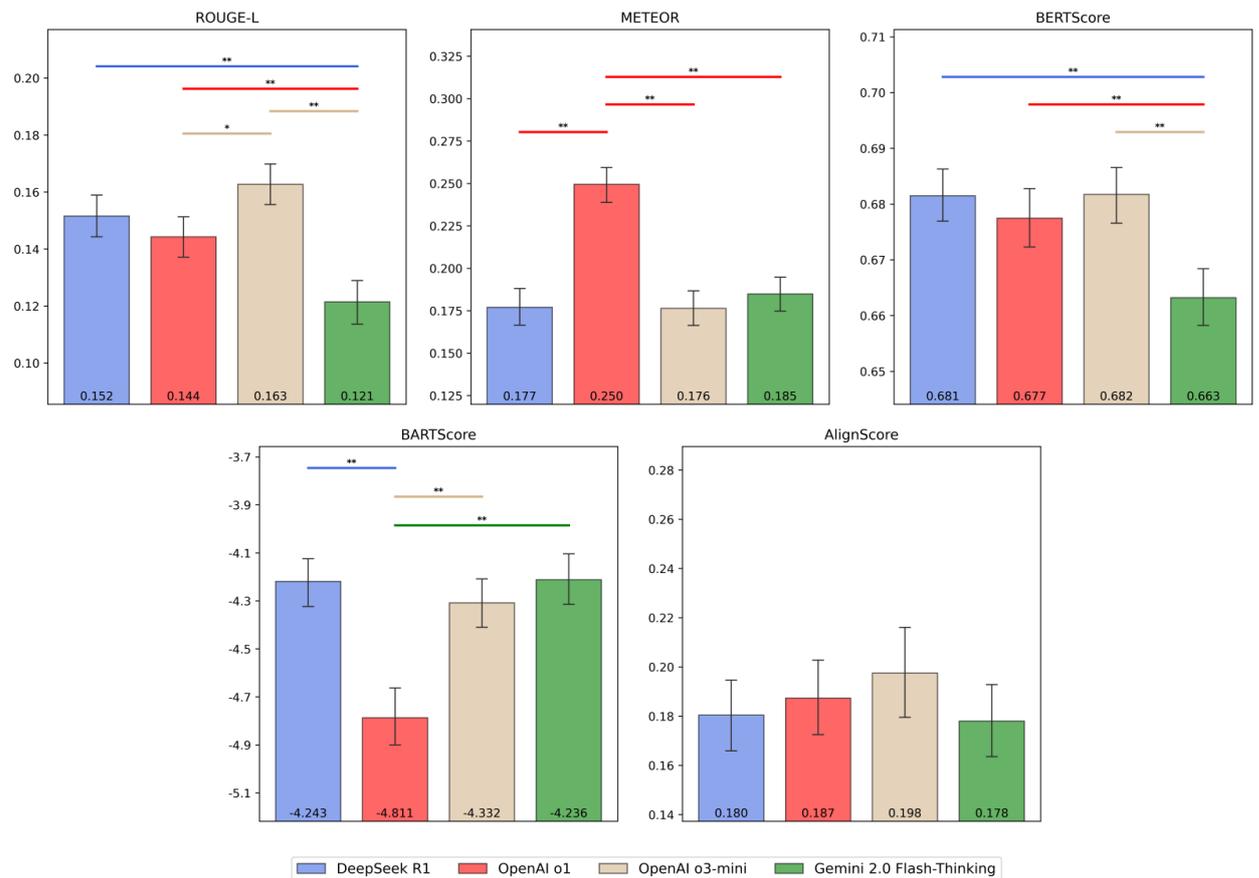

Horizontal bars indicate statistically significant differences between pairs of LLMs, with the color of the bar representing the LLM that performed better. Error bars for each metric indicate 95% confidence intervals obtained from bootstrapping. * denotes $p < 0·05$, ** denotes $p < 0·001$, both after Bonferroni correction.

Supplementary Table 1: Definitions of the Text-Generation Metrics Used in the Study.

| Text Generation Metric | Focus | Sensitive To | Example Comparison |
|---|---|---|---|
| ROUGE-L | Exact word order in the longest common sequence between the correct answer and the model output | Word order, exact lexical matches, length of matching sequences, does not account for meaning or context, surface-level word forms | Example correct answer: "The patient has diabetic retinopathy." High scoring model output: "the patient has diabetic retinopathy." Low scoring model output: "diabetic retinopathy affects the patient." (word order differs) |
| METEOR | Unigram matching the surface forms, stemmed forms, synonyms and morphology | Word matches (exact or synonym), word order (but more flexible than ROUGE) stemming matches (e.g., "running" and "runs" count as matches), has | Example correct answer: "The patient exhibits severe vision loss." High scoring model output: "the patient shows significant vision loss." (synonyms and similar wording) Low scoring model output: "the patient demonstrates profound blindness." (meaning shift, |

| | | limited semantic understanding, surface-level word forms | blindness ¹ vision loss) |
|---|---|---|---|
| BERTScore | Contextual semantic similarity | Contextual meaning, token-level alignments, less affected by exact word choice, can understand paraphrases | Example correct answer: "The individual is experiencing blurred vision due to diabetes." High scoring model output: "Diabetes is causing the patient's vision to blur." (different words, same meaning) Low scoring model output: "the individual is experiencing high blood pressure." (meaning is unrelated despite similar sentence form) |
| BARTScore | Contextual semantic similarity, fluency and probability of generating the candidate text | Contextual meaning, fluency of output, less affected by surface-level word forms, | Example correct answer: "this patient presents with symptoms of glaucoma." High scoring model output: "symptoms of glaucoma are what this patient presents with." (slightly |
| | given the reference (or reference given candidate) | | awkward but fluent) Low scoring model output: "Patient glaucoma symptoms." (broken grammar, low fluency) |
| AlignScore | Factual consistency (where fact A in the correct answer is contained in the model output), semantic alignment | Content selection and preservation (Whether fact A in the correct answer is contained in the model output), whether key idea in correct answer is in the model output | Example correct answer: "early detection of macular degeneration is essential to prevent vision loss." High scoring model output: "to avoid vision loss, it's crucial to detect macular degeneration early." (same key ideas but reordered) Low scoring model output: "macular degeneration is a disease." (not fatally aligned) |

Supplementary Table 2. Comparative Performance of DeepSeek-R1, OpenAI o1, o3-mini, and Gemini 2.0 Flash-Thinking on the MedMCQA ophthalmology subset with the "Cataract" sub-specialty (n = 330).

| Metric | DeepSeek R1 | OpenAI o1 | OpenAI o3-mini | Gemini 2.0 Flash-Thinking |
|---|---|---|---|---|
| Accuracy | 0·887 | **0·916** | 0·858 | 0·790 |
| Macro-F1 | 0·886 (0·845 - 0·922) | **0·914 (0·881 - 0·944)** | 0·852 (0·811 - 0·892) | 0·785 (0·738 - 0·830) |
| ROUGE-L | 0·148 (0·142 - 0·153) | 0·140 (0·134 - 0·146) | **0·158 (0·152 - 0·164)** | 0·121 (0·115 - 0·127) |
| METEOR | 0·167 (0·159 - 0·175) | **0·233 (0·225 - 0·242)** | 0·168 (0·159 - 0·177) | 0·168 (0·161 - 0·176) |
| BERTScore | **0·677 (0·673 - 0·681)** | 0·674 (0·670 - 0·679) | 0·675 (0·671 - 0·680) | 0·660 (0·655 - 0·664) |
| BARTScore | **-4·281 (-4·368 - -4·199)** | -4·895 (-5·005 - -4·794) | -4·354 (-4·444 - -4·268) | -4·320 (-4·412 - -4·228) |
| AlignScore | 0·187 (0·174 - 0·199) | 0·199 (0·186 - 0·214) | **0·205 (0·190 - 0·220)** | 0·169 (0·158 - 0·180) |

Supplementary Table 3. Comparative Performance of DeepSeek-R1, OpenAI o1, o3-mini, and Gemini 2.0 Flash-Thinking on the MedMCQA ophthalmology subset with the "Retinal Diseases" sub-specialty (n = 307).

| Metric | DeepSeek R1 | OpenAI o1 | OpenAI o3-mini | Gemini 2.0 Flash-Thinking |
|---|---|---|---|---|
| Accuracy | 0·888 | **0·902** | 0·902 | 0·835 |
| Macro-F1 | 0·879 (0·834 - 0·918) | **0·897 (0·861 - 0·932)** | 0·896 (0·857 - 0·931) | 0·832 (0·787 - 0·878) |
| ROUGE-L | 0·163 (0·156 - 0·169) | 0·153 (0·146 - 0·159) | **0·172 (0·165 - 0·178)** | 0·130 (0·124 - 0·136) |
| METEOR | 0·183 (0·176 - 0·190) | **0·251 (0·241 - 0·260)** | 0·182 (0·173 - 0·191) | 0·195 (0·187 - 0·203) |
| BERTScore | **0·689 (0·685 - 0·693)** | 0·684 (0·680 - 0·689) | 0·688 (0·684 - 0·692) | 0·673 (0·668 - 0·677) |
| BARTScore | **-4·112 (-4·201 - -4·031)** | -4·698 (-4·796 - -4·603) | -4·192 (-4·281 - -4·107) | -4·135 (-4·227 - -4·041) |
| AlignScore | 0·188 (0·175 - 0·202) | 0·204 (0·190 - 0·219) | **0·216 (0·201 - 0·232)** | 0·182 (0·170 - 0·194) |

Supplementary Table 4. Comparative Performance of DeepSeek-R1, OpenAI o1, o3-mini, and Gemini 2.0 Flash-Thinking on the MedMCQA ophthalmology subset with the "Glaucoma" sub-specialty (n = 208).

| Metric | DeepSeek R1 | OpenAI o1 | OpenAI o3-mini | Gemini 2.0 Flash-Thinking |
|---|---|---|---|---|
| Accuracy | 0·865 | **0·896** | 0·870 | 0·833 |
| Macro-F1 | 0·860 (0·807 - 0·909) | **0·893 (0·846 - 0·936)** | 0·865 (0·812 - 0·912) | 0·827 (0·775 - 0·877) |
| ROUGE-L | 0·152 (0·144 - 0·159) | 0·144 (0·137 - 0·151) | **0·163 (0·156 - 0·170)** | 0·121 (0·114 - 0·129) |
| METEOR | 0·177 (0·166 - 0·188) | **0·249 (0·239 - 0·259)** | 0·177 (0·166 - 0·187) | 0·185 (0·175 - 0·195) |
| BERTScore | 0·682 (0·677 - 0·686) | 0·677 (0·672 - 0·683) | **0·682 (0·677 - 0·687)** | 0·663 (0·658 - 0·668) |
| BARTScore | -4·244 (-4·347 - -4·148) | -4·811 (-4·924 - -4·687) | -4·331 (-4·434 - -4·232) | **-4·235 (-4·338 - -4·128)** |
| AlignScore | 0·180 (0·166 - 0·195) | 0·188 (0·172 - 0·203) | **0·197 (0·180 - 0·216)** | 0·178 (0·164 - 0·193) |

Supplementary Table 5. Thinking Processes and Final Output Reasonings of the Four Reasoning-Focused LLMs for an input question from the "Cataract" sub-specialty of the MedMCQA ophthalmology dataset.

Question: An infant presents with bilateral white pupillary reflex. On slit lamp examination a zone of opacity is observed around the fetal nucleus with spoke like radial opacities. The most likely diagnosis is:

Options: A: Cataracta Centralis Pulverulenta, **B: Lamellar cataract (correct)**, C: Coronary cataract, D: Posterior polar cataract

| **(A) DeepSeek-R1** |
|---|
| **Thinking process:** <br><br> Okay, let's tackle this ophthalmology question. The scenario is an infant with bilateral white pupillary reflex, and on slit lamp exam, there's a zone of opacity around the fetal nucleus with spoke-like radial opacities. The options are four types of cataracts. <br><br> First, I need to recall the different types of cataracts mentioned here. Let's break them down. <br><br> Posterior polar cataract (D) is usually a small, central opacity at the posterior pole. It might be associated with persistent fetal vasculature but I don't think it has spokes. It's more like a dense spot in the back. |

Lamellar cataract (B) is a type where there's a layer of opacity surrounding a clearer nucleus, like a shell. It's also called zonular cataract. The opacity is around the embryonic nucleus, which is the central part. Since the question mentions the fetal nucleus, which is a bit later in development than the embryonic nucleus, maybe lamellar is around the embryonic. The spoke-like opacities don't ring a bell here, more like a disc-shaped opacity.

Coronary cataract (C) is characterized by club-shaped opacities arranged around the equator like a crown. They're in the cortex, and might look like spokes when viewed from the front. Coronary cataracts are often present at birth but might develop later. The description of spoke-like radial opacities around the fetal nucleus might fit here. Wait, the question says the opacity is around the fetal nucleus, so maybe the spokes are extending out from there.

Cataracta Centralis Pulverulenta (A) is a central dusty opacity in the nucleus. It's a congenital cataract, appearing as a small, powdery central opacity. This might be a nuclear cataract, but the question mentions opacity around the fetal nucleus, not in the center. Also, the spoke-like aspect isn't part of this description. So probably not A.

So the key points: around the fetal nucleus (so surrounding it, not central), with spoke-like opacities. Lamellar (B) has a layer surrounding the nucleus, which is the embryonic nucleus. The fetal nucleus is formed later. If the opacity is around the fetal nucleus, perhaps that's a different layer. Coronary (C) has spokes in the cortex, arranged radially. If they are around the equator, maybe the spokes are in the periphery, but the question says around the fetal nucleus.

Alternatively, if the spokes are in the area surrounding the fetal nucleus, which is the deeper layers.

Wait, another thought: Lamellar cataracts affect the lamellar zone, which is between the embryonic and fetal nuclei. So the opacity is in that zone. If the opacities are surrounding the fetal nucleus, that could be lamellar. But the spoke-like opacities - lamellar usually forms a disc, not spokes. Coronary cataracts have radial opacities, like spokes, which could look like a crown. So if it's around the fetal nucleus and has spokes, maybe Coronary (C) is the answer. But I'm a bit confused here.

Alternatively, the description might match a sutural cataract, but that's not one of the options. Sutural cataracts are along the Y-sutures of the lens. But the options given don't include that.

Wait, another angle: The fetal nucleus is the part of the lens formed during the fetal period. The embryonic nucleus is the earliest part. Lamellar cataract affects the layer around the embryonic nucleus, so it's a zonular opacity surrounding the embryonic nucleus. If the opacity here is around the fetal nucleus, that might be a different zone. Maybe the "spoke-like radial opacities" are in the region around the fetal nucleus. The options are B, C... Let me check standard sources.

From what I remember, lamellar cataracts present as a cloudy layer surrounding a clear nucleus, creating a kind of shell. They can have radial opacities extending outward like spokes from the central shell, which is called "riders" or spokes. These riders might extend into the cortex. So the combination of a central shell-like opacity (around the nucleus) and spokes would be lamellar. The description here is a zone of opacity around the fetal nucleus with spokes. So that fits lamellar (B).

Alternatively, Coronary cataracts (C) have club-shaped opacities in the cortex, arranged circularly, which might look like a crown when viewed from the front. These are in the periphery, perhaps not around the fetal nucleus. The opacities in coronary are more cortical and maybe not involving the fetal nucleus.

So, putting it together, the correct answer would be B: Lamellar cataract. Because lamellar cataract involves an opaque layer surrounding the nucleus (fetal in this case), and can have radial spoke-like opacities extending from it. The central pulverulenta is too central, and the others don't match the spoke appearance. Posterior polar is at the back.

**Output response:**

The infant's presentation of bilateral white pupillary reflex and slit lamp findings of a zone of opacity around the fetal nucleus with spoke-like radial opacities align with the features of a lamellar cataract. Lamellar (zonular) cataracts are characterized by a cloudy layer (opacity) surrounding a clearer nucleus, often affecting the zone between the embryonic and fetal nuclei. These cataracts may exhibit radial spoke-like extensions ('riders') that project into the surrounding cortex, explaining the observed radial opacities. Option A (Cataracta Centralis Pulverulenta) describes a central powdery nuclear opacity, which does not involve surrounding zones or spokes. Option C (Coronary cataract) presents with peripheral club-shaped cortical opacities arranged coronally, unrelated to the fetal nucleus. Option D (Posterior polar cataract) involves a dense central posterior capsule opacity, lacking radial spokes. Thus, lamellar cataract (B) is the most consistent diagnosis based on the anatomical location and morphological features described.

**(B) OpenAI o1**

**Thinking Process:**

Assessing the diagnosis

I'm taking a closer look at the infant's bilateral white pupillary reflex and examining the slit-lamp exam findings. The zones of opacity and spoke-like radial opacities suggest a lamellar cataract as the likely diagnosis.

Identifying lamellar cataract

I'm piecing together the typical presentation of a lamellar cataract: opacification around the fetal nucleus with clear central portions and radial opacities, suggesting the correct diagnosis.

Ruling out possibilities

I'm concluding that coronary cataract, with its peripheral cortex opacities typical in older individuals, and posterior polar cataract, with opacities behind the fetal nucleus, are unlikely given the bilateral white pupillary reflex.

Determining the cause

I'm concluding that the lamellar cataract diagnosis is confirmed by the oval ring with spoke-like radial opacities around the fetal nucleus observed during the slit-lamp examination.

**Output Response:**

The most likely diagnosis is Lamellar cataract (Option B). Here's the reasoning:

1. Bilateral White Pupillary Reflex (Leukocoria): The infant presents with bilateral white pupillary reflexes, indicating the presence of opacities in the lens of both eyes. This is commonly seen in congenital cataracts.

2. Zone of Opacity Around the Fetal Nucleus: Lamellar cataracts are characterized by opacities that surround the embryonic or fetal nucleus of the lens. They involve a particular lamella or layer of the lens while other layers remain clear.

3. Spoke-like Radial Opacities: Lamellar cataracts often have spoke-like opacities that extend radially. These cortical riders are typical features of lamellar cataracts and result from developmental anomalies during lens fiber formation.

4. Age of Presentation: Lamellar cataracts are usually congenital or can develop in early childhood, fitting the infant's age in this case.

5. Exclusion of Other Options:

   - Option A (Cataracto Centralis Pulverulenta): This type of cataract involves powder-like opacities in the central nucleus but doesn't typically present with spoke-like radial opacities or involve the surrounding lamella.

   - Option C (Coronary Cataract): Coronary cataracts are characterized by club-shaped opacities in the cortex near the equator of the lens, resembling a crown. They are peripheral and do not involve the fetal nucleus or present with a white pupillary reflex.

   - Option D (Posterior Polar Cataract): This cataract occurs at the posterior pole of the lens and appears as a central opacity on the posterior capsule. It doesn't present with radial spokes or a zonal opacity around the fetal nucleus.

In conclusion, the presentation of bilateral white pupillary reflexes in an infant, along with a zonal opacity around the fetal nucleus and spoke-like radial opacities, is most consistent with a Lamellar cataract.

**(C) OpenAI o3-mini**

**Thinking Process:**

I'm analyzing the situation and narrowing down the options. The baby has a bilateral white pupillary reflex and, on slit lamp exam, a zone of opacity around the fetal nucleus with radial "spoke-like" opacities. This pattern matches lamellar cataract (also called zonular cataract). It involves the fetal nucleus and spares the embryonic nucleus. The distinct "spoke-like" appearance is a key feature. So, the most consistent diagnosis here is lamellar cataract, answer B. I'll need to present a full, step-by-step ophthalmology explanation.

**Output Response:**

Lamellar cataracts are characterized by an opacity that surrounds the fetal nucleus, often presenting with a zone of opacity and spoke-like radial opacities. In this case, the slit lamp examination showing a zone of opacity around the fetal nucleus along with radial opacities is classic for lamellar cataract. Other options such as Cataracto Centralis Pulverulenta, Coronary cataract, and Posterior polar cataract have different clinical presentations; for example, posterior polar cataracts affect the posterior part of the lens, while coronary cataracts typically display opacities in the periphery arranged like a crown. Therefore, the correct diagnosis based on the given examination findings is lamellar cataract.

**(D) Gemini 2.0 Flash-Thinking**

**Thinking Process:**

The user is asking for the diagnosis of a congenital cataract based on the description of its morphology in an infant presenting with leukocoria (white pupillary reflex).

Let's break down the key features from the question:

Bilateral White Pupillary Reflex (Leukocoria): This is the primary presenting symptom, strongly suggesting a condition obstructing the visual axis in both eyes, most likely cataracts in infants. Leukocoria in infants raises immediate concern for conditions like retinoblastoma, but in this context, given the options provided are all cataract types, we should focus on cataracts.

Infant: Age is crucial. Congenital cataracts are cataracts present at birth or developing shortly after.

Slit Lamp Examination Findings:

Zone of opacity around the fetal nucleus: This localizes the opacity within the lens. The fetal nucleus is the central part of the lens formed during fetal development.

Spoke-like radial opacities: This is the key descriptive feature of the opacity's morphology. "Spoke-like" and "radial" suggest opacities extending outwards from the central zone, resembling spokes of a wheel.

Now let's analyze each option and see which one fits best with these features:

Option A: Cataracta Centralis Pulverulenta: "Pulverulenta" implies a dusty or powdery appearance. "Centralis" indicates central location. While it's central, the description of "spoke-like radial opacities" is less typical for a "dusty" central cataract. However, the central location is a point in favor.

Option B: Lamellar cataract: Lamellar cataracts, also known as zonular cataracts, are a very common type of congenital cataract. They are characterized by opacities in a specific lamella or layer of the lens, often surrounding a clear central nucleus and cortex. The "zone of opacity around the fetal nucleus" and "spoke-like radial opacities" strongly resonate with the description of lamellar cataracts. They often have these radiating opacities extending outwards in a spoke-like fashion from the affected lamella. This is a very strong contender.

Option C: Coronary cataract: Coronary cataracts are typically found in the periphery of the lens cortex, resembling a "crown" or "corona". The description of "opacity around the fetal nucleus" goes against coronary cataracts which are peripheral. Coronary cataracts also typically present later in childhood or even adolescence and are often blue-dot cataracts arranged in a ring-like fashion in the periphery. This option seems less likely due to location.

Option D: Posterior polar cataract: Posterior polar cataracts are located at the posterior pole of the lens, right at the back. The description "zone of opacity around the fetal nucleus" suggests a more central location rather than specifically posterior polar. While they can cause leukocoria, their morphology is typically a plaque-like or disc-shaped opacity at the posterior lens capsule, not necessarily "spoke-like radial opacities" around the fetal nucleus. This seems less likely due to location and morphology description.

Comparing Option A and Option B in detail:

Cataracta Centralis Pulverulenta: More of a central dusty opacity, less described with prominent "spoke-like radial opacities". While location is central, the morphology fit isn't perfect.

Lamellar cataract: Classically described with opacities in a lamella surrounding the nucleus and very often described with spoke-like or rider-like opacities radiating outwards from the affected lamella. Location (around fetal nucleus) and morphology (spoke-like) fit very well. Lamellar cataracts are also among the most common congenital cataracts presenting with bilateral leukocoria.

Conclusion: Lamellar cataract fits almost perfectly with all aspects of the description – bilateral leukocoria in an infant, zone of opacity around the fetal nucleus, and crucially, the spoke-like radial opacities.

Therefore, Lamellar cataract (Option B) is the most likely diagnosis.

**Output Response:**

The question describes an infant presenting with bilateral leukocoria (white pupillary reflex). This immediately points towards a condition obstructing light from reaching the retina, and in infants, congenital cataracts are a common cause. The slit lamp examination further details a 'zone of opacity around the fetal nucleus with spoke like radial opacities'. This specific description is highly characteristic of Cataracto Centralis Pulverulenta (A), also known as a pulverulent nuclear cataract.

Here's why option A is the most likely and why the others are less likely:

A: Cataracto Centralis Pulverulenta: This type of congenital cataract is specifically characterized by a central, often dust-like or pulverulent (hence the name) opacity around the fetal nucleus. The 'spoke-like radial opacities' are also a common feature, resembling fine radiating lines from the central opacity. This description perfectly matches the clinical findings in the question. Pulverulent cataracts are often bilateral and can present with leukocoria in infancy, aligning with the patient's presentation.

B: Lamellar cataract: Lamellar cataracts (also known as zonular cataracts) are characterized by an opacity in a layer (lamella or zone) of the lens, often surrounding a clear central nucleus and periphery. While they can be bilateral and cause leukocoria, the description of 'opacity *around* the fetal nucleus with spoke-like *radial* opacities' is not the typical presentation of a lamellar cataract. Lamellar cataracts are more often described as having a ring-shaped opacity or onion-skin appearance rather than radial spokes emanating from the central nucleus.

C: Coronary cataract: Coronary cataracts are typically located in the periphery of the lens cortex, appearing as bluish or whitish opacities in a crown-like pattern around the periphery. They are usually found in adults and older children and are less likely to be a cause of leukocoria in infants because they are peripheral and do not significantly obstruct the central visual axis. Furthermore, the description of 'opacity around the fetal nucleus' directly contradicts the peripheral location of coronary cataracts.

D: Posterior polar cataract: Posterior polar cataracts are located at the posterior capsule of the lens. They are characterized by a plaque-like or disc-shaped opacity at the back of the lens. While they can cause leukocoria, the description of 'opacity around the fetal nucleus with spoke-like radial opacities' is not typical for posterior polar cataracts. Posterior polar cataracts are not primarily described as being 'around the fetal nucleus' nor are they typically associated with spoke-like radial opacities emanating from a central zone.

In summary: The combination of bilateral leukocoria, central opacity around the fetal nucleus, and spoke-like radial opacities points most strongly towards Cataracto Centralis Pulverulenta. The other options, while types of cataracts, do not fit the specific descriptive details provided in the question as closely as Cataracto Centralis Pulverulenta does.

Supplementary Table 6. Thinking Processes and Final Output Reasonings of the Four Reasoning-Focused LLMs for an input question from the "Retinal Diseases" sub-specialty of the MedMCQA ophthalmology dataset.

Question: A young patient developed sudden painless loss of vision which recovers spontaneously over a period of 3 months. What is the most probable diagnosis?

Options: A: Macular hole, B: Angle closure glaucoma, C: Retinal detachment, **D: Central Serous retinopathy (correct)**

| **(A) DeepSeek-R1** |
| --- |
| **Thinking process:** |
| Okay, let's see. The question is about a young patient with sudden painless vision loss that recovers on its own over three months. The options are Macular hole, Angle closure glaucoma, Retinal detachment, and Central Serous retinopathy. |
| First, I should think about each condition. Macular hole - I know that causes vision loss, typically central, and it can be sudden. But does it resolve on its own? I think macular holes usually don't resolve without surgery. They might progress, so maybe not A here. |
| Angle closure glaucoma (B) is an emergency with painful vision loss, high IOP, red eye. But the question states the vision loss is painless. So B is probably out. |

Retinal detachment (C) presents with flashes, floaters, a curtain over vision. It's also typically urgent and doesn't resolve without treatment. If untreated, vision loss is permanent, so spontaneous recovery is unlikely. So maybe not C either.

Central Serous retinopathy (D) comes to mind. It's more common in young to middle-aged adults, especially men. Symptoms include sudden blurred vision, sometimes with a central blind spot. It's characterized by fluid accumulation under the retina, leading to serous detachment. Importantly, many cases resolve spontaneously over weeks to months. That fits the scenario here: painless, sudden, and resolves in 3 months. So D seems like the right answer.

**Output Response:**

Central Serous Retinopathy (CSR) is the most likely diagnosis in this scenario. CSR typically presents in young to middle-aged adults (often males) with sudden, painless vision loss or distortion, often resolving spontaneously within 3-6 months. The condition involves fluid accumulation under the retina (subretinal fluid), leading to a localized serous detachment. Spontaneous resolution occurs as the fluid reabsorbs. In contrast, a macular hole (A) causes central vision loss but rarely resolves without intervention. Angle closure glaucoma (B) presents with painful vision loss due to elevated intraocular pressure. Retinal detachment (C) causes symptoms like flashes, floaters, and a 'curtain-like' vision loss, which is irreversible without surgical treatment. Thus, the clinical features (sudden, painless, self-resolving) align best with CSR.

**(B) OpenAI o1**

**Thinking Process:**

Assessing probable diagnosis

I'm breaking down symptoms of sudden painless vision loss in a young patient, trying to link the timeline and common causes, leading me to central serous retinopathy.

Pinpointing the condition

Interestingly enough, central serous retinopathy's hallmark is a retinal detachment, often in the macular area. It recovers spontaneously in 3 months, which aligns perfectly with the given symptoms.

**Response:**

Central Serous Retinopathy (CSR) is most commonly seen in young to middle-aged adults, particularly males. It presents with sudden, painless loss of central vision due to the accumulation of serous fluid under the macula, leading to a localized detachment of the neurosensory retina. A hallmark of CSR is its tendency to resolve spontaneously, with vision often recovering over a period of weeks to months, typically within 3 months, as the subretinal fluid reabsorbs and the retina reattaches. Therefore, in a young patient who experiences sudden painless loss of vision that recovers spontaneously over three months, CSR is the most probable diagnosis.

In contrast:

- Option A: Macular hole usually presents with a gradual decrease in central vision and metamorphopsia (distorted vision). Spontaneous closure is rare, and surgical intervention is often required for improvement.

- Option B: Angle closure glaucoma is characterized by sudden loss of vision accompanied by severe eye pain, headache, nausea, vomiting, and halos around lights due to a rapid increase in intraocular pressure. It is a painful condition, which does not align with the painless presentation.

- Option C: Retinal detachment presents with sudden, painless loss of vision but does not resolve spontaneously. Patients may report flashes of light, floaters, or a curtain-like shadow over their visual field. Prompt surgical intervention is necessary to reattach the retina and restore vision.

Therefore, considering the age, sudden painless vision loss, and spontaneous recovery over three months, Central Serous Retinopathy is the most consistent diagnosis.

| **(C) OpenAI o3-mini** |
|---|
| **Thinking Process:**<br><br>Let's break this down:<br><br>Macular hole typically affects older people and doesn't usually recover spontaneously — not the right choice.<br><br>Angle closure glaucoma is painful with red eyes, so not it.<br><br>Retinal detachment is also typically painful, progressive, and doesn't resolve on its own, so that's out too.<br><br>Central serous retinopathy (CSR) is common in younger patients, can cause sudden painless vision loss, and typically recovers on its own in a few months — that's the most fitting option. |
| **Response:**<br><br>Central Serous Retinopathy (CSR) typically affects young to middle-aged adults and presents with a sudden, painless loss of vision, often characterized by central visual distortion. One of the hallmark features of CSR is its dynamic course; the condition usually resolves spontaneously over weeks to a few months (approximately 3 months), with the fluid under the retina gradually reabsorbing. In contrast, the other options do not fit this clinical scenario: a |

macular hole usually occurs in older individuals and does not resolve spontaneously; angle closure glaucoma presents with severe pain and acute symptoms; and retinal detachment, though painless, does not typically resolve without intervention and may lead to permanent vision loss. Therefore, the most probable diagnosis is Central Serous Retinopathy.

**(D) Gemini 2.0 Flash-Thinking**

**Thinking Process:**

The user is asking for the most probable diagnosis for a young patient experiencing sudden, painless vision loss that recovers spontaneously in 3 months. I need to evaluate each option based on these key characteristics: sudden onset, painless, spontaneous recovery, and typical patient demographics.

Let's break down each option and see how well it fits the scenario.

Sudden painless loss of vision: This suggests a vascular, neurological, or retinal event, rather than an inflammatory or traumatic process that would likely cause pain. It also points away from slow progressive conditions.

Spontaneous recovery: This is a crucial point. Many ophthalmic conditions cause permanent vision loss unless treated. Spontaneous recovery strongly suggests a condition that is self-limiting or resolves naturally.

Over 3 months: This timeframe indicates a condition that is not immediately transient (like amaurosis fugax which recovers within minutes) but also not chronic or irreversible.

Young patient: While not specifically defined, "young" generally means not elderly, potentially ruling out some age-related degenerative conditions as the primary suspect, although they are still possible.

Now let's evaluate each option:

A: Macular hole: Macular holes typically cause gradual vision loss, although it can sometimes be perceived as sudden if a smaller hole suddenly enlarges. Macular holes are rarely painless. Crucially, macular holes do not spontaneously recover vision to a significant degree. They usually require surgical intervention (vitrectomy) for visual improvement. This option is unlikely due to lack of spontaneous recovery and typical gradual onset.

B: Angle closure glaucoma: Acute angle closure glaucoma presents with sudden vision loss, but it is characteristically painful. The pain is often severe, accompanied by headache, nausea, vomiting, and halos around lights. While vision can be restored if treated promptly, spontaneous recovery in untreated acute angle closure glaucoma is not expected. Untreated acute angle closure can lead to permanent optic nerve damage and vision loss. Chronic angle closure is typically more gradual and may be painless in the early stages, but sudden complete painless loss is not typical, nor is spontaneous recovery over 3 months without intervention in established cases. This option is unlikely due to the "painless" aspect and lack of spontaneous recovery in untreated acute cases.

C: Retinal detachment: Retinal detachment can present with sudden vision loss, often described as a curtain coming down over the vision or sudden onset of floaters and flashes. Retinal detachment is painless. However, spontaneous recovery from a retinal detachment is extremely rare. Retinal detachment typically requires urgent surgical intervention to reattach the retina and prevent permanent vision loss. Untreated retinal detachment will generally lead to permanent vision loss, not spontaneous recovery. This option is highly unlikely due to the lack of spontaneous recovery.

D: Central Serous Retinopathy (CSR): Central Serous Retinopathy (CSR) is a condition where fluid accumulates under the retina in the macula, often causing sudden onset of blurred or distorted vision. Vision loss in CSR can sometimes be described as sudden by patients when the central vision is acutely affected. CSR is characteristically painless. And importantly, CSR often resolves spontaneously over weeks to months (typically within 1-3 months). Young to middle-aged males are a common demographic for CSR, although it can occur in anyone. The visual recovery in CSR is due to the spontaneous reabsorption of subretinal fluid. This option fits all the key characteristics: sudden onset, painless, spontaneous recovery in 3 months, and common in younger individuals.

Conclusion:

Based on the characteristics provided: sudden painless vision loss and spontaneous recovery within 3 months in a young patient, Central Serous Retinopathy (CSR) is the most probable diagnosis among the given options.

Final Answer: D: Central Serous retinopathy

**Output Response:**

The most probable diagnosis is Central Serous Retinopathy (CSR), often characterized by sudden, painless vision loss that improves spontaneously. Here's the breakdown:

Central Serous Retinopathy (CSR): This condition involves fluid accumulation under the retina, typically in the macula, causing blurred or distorted vision, sometimes with a central scotoma (blind spot). It often affects young to middle-aged adults, and typically resolves spontaneously within 3-6 months. The vision loss is often described as painless and may include changes in color perception and metamorphopsia (distorted vision). The recovery pattern described in the question perfectly aligns with the typical course of CSR.

Macular Hole (A): A macular hole causes a permanent central vision loss, not spontaneous recovery. Visual acuity is permanently and severely reduced, and does not spontaneously recover.

Angle Closure Glaucoma (B): Angle closure glaucoma presents with acute, painful vision loss accompanied by symptoms like headache, nausea, eye pain, and blurred vision, often with halos around lights. It does not recover spontaneously; it requires immediate intervention to lower intraocular pressure and prevent blindness. The painless, spontaneous recovery is not consistent with glaucoma.

Retinal Detachment (C): Retinal detachment presents with sudden painless vision loss, however, it does not have spontaneous recovery. It can present with a curtain-like shadow across the visual field, floaters, and flashes of light (photopsia). Without prompt surgical intervention, it leads to permanent vision loss. Although the vision loss can be painless, the lack of spontaneous recovery differentiates it from the case described.

Supplementary Table 7. Thinking Processes and Final Output Reasonings of the Four Reasoning-Focused LLMs for an input question from the "Glaucoma" sub-specialty of the MedMCQA ophthalmology dataset.

Question: A 70-year-old patient presents with progressive deterioration of vision. On examination, the pupillary reaction is sluggish and the IOP is normal. Fundoscopy shows a large and deep cup. Visual field reveals paracentral scotoma. What is the probable diagnosis

Options: **A: Normal tension glaucoma (correct)**, B: Absolute glaucoma, C: Primary angle closure glaucoma, D: Neovascular glaucoma

| (A) DeepSeek-R1 |
| --- |
| **Thinking process:** <br><br> Okay, let's try to figure this out. The patient is 70 with progressive vision loss. The key findings here are the sluggish pupillary reaction, normal IOP, a large and deep cup on fundoscopy, and paracentral scotoma in the visual field. Let's break this down. <br><br> First, the normal intraocular pressure (IOP) is important because it points away from conditions where IOP is typically high. Primary angle closure glaucoma (option C) usually presents with elevated IOP, so that might not fit here. Neovascular glaucoma (option D) is associated with new blood vessel growth, often due to diabetes or retinal vein occlusion, and IOP is usually high here as well. Absolute glaucoma (option B) is end-stage with very high IOP and no light |

perception, which isn't mentioned here.

Normal tension glaucoma (option A) is when there's optic nerve damage and visual field loss despite IOP being within normal range. The findings of a large cup and paracentral scotoma align with glaucoma, especially since the fields show paracentral loss, which can be early in NTG. The sluggish pupil reaction could be due to the optic nerve damage affecting the afferent pathway. So putting it all together, the correct answer seems to be A.

**Output Response:**

The patient's presentation aligns with normal tension glaucoma (NTG). Key indicators include normal intraocular pressure (IOP), a large and deep optic cup on fundoscopy, and paracentral scotoma on visual field testing. NTG is characterized by glaucomatous optic neuropathy (e.g., increased cup-to-disc ratio) and visual field defects despite IOP within the statistically normal range. The sluggish pupillary reaction reflects damage to the afferent pathway due to optic nerve atrophy. Paracentral scotomas are commonly seen early in NTG progression. Absolute glaucoma (B) involves complete vision loss and elevated IOP. Primary angle closure glaucoma (C) typically presents with elevated IOP and a closed anterior chamber angle. Neovascular glaucoma (D) is associated with elevated IOP and retinal ischemia, none of which are present here.

**(B) OpenAI o1**

**Thinking Process:**

Evaluating clinical signs

I'm piecing together the patient's age and progressive vision loss. His sluggish pupillary reaction aligns with past symptoms. Normal IOP and a large, deep cup suggest normal tension glaucoma.

**Output Response:**

The patient is a 70-year-old with progressive deterioration of vision, a sluggish pupillary reaction, normal intraocular pressure (IOP), a large and deep optic disc cup observed on fundoscopy, and a paracentral scotoma on visual field testing. These findings are characteristic of normal tension glaucoma (NTG).

Explanation:

- Normal Tension Glaucoma (NTG): NTG is a form of open-angle glaucoma where optic nerve damage occurs despite IOP being within the normal range. It is more common in older adults.
  - Optic Disc Cupping: The large and deep cup indicates glaucomatous optic neuropathy, a hallmark of glaucoma, including NTG.
  - Visual Field Defects: Paracentral scotomas are common early visual field defects in NTG,

reflecting damage to the retinal nerve fibers near the macula.

- Sluggish Pupillary Reaction: This suggests optic nerve dysfunction, consistent with glaucomatous damage.

- Exclusion of Other Options:

  - Option B (Absolute Glaucoma): This is an end-stage glaucoma where the eye is blind and often painful, with a fixed dilated pupil and very high IOP. The patient still has vision and normal IOP, making this less likely.

  - Option C (Primary Angle Closure Glaucoma): Typically presents acutely with severe eye pain, headache, nausea, vomiting, markedly elevated IOP, a mid-dilated fixed pupil, and corneal edema not the case here.

  - Option D (Neovascular Glaucoma): Occurs secondary to retinal ischemia (e.g., diabetic retinopathy), leading to neovascularization of the iris and angle structures, causing elevated IOP and pain. The patient does not present with high IOP or pain.

Therefore, the most probable diagnosis is normal tension glaucoma (Option A).

**(C) OpenAI o3-mini**

**Thinking Process:**

The user asks for my reasoning and thought process, but the guidelines say that I shouldn't reveal the internal chain-of-thought. I'm supposed to summarize the reasoning behind the decisions I make without showing the step-by-step details from my internal process.

So, I'll explain my reasoning in a simplified manner, focusing on the steps and summary conclusions, without diving into the behind-the-scenes chain-of-thought thinking directly. This should strike the right balance!

**Output Response:**

The patient presents with progressive deterioration of vision along with a deep and large cup (a classic sign of glaucomatous optic neuropathy) and a paracentral scotoma, which are common features of glaucoma. Despite these glaucomatous changes, the intraocular pressure (IOP) is normal, thereby ruling out other forms of glaucoma where IOP would typically be elevated (such as primary angle closure glaucoma or neovascular glaucoma). Absolute glaucoma is also excluded since it typically represents an end-stage glaucoma with no remaining vision, which does not fit the scenario described. Therefore, the most likely diagnosis is Normal Tension Glaucoma, which is characterized by glaucomatous optic nerve damage and typical visual field defects in the presence of normal IOP.

**(D) Gemini 2.0 Flash-Thinking**

**Thinking Process:**

The user is asking for a diagnosis based on the provided clinical information. The patient is 70 years old and has progressive vision deterioration. The key findings are:

Sluggish pupillary reaction: This indicates a problem with the pupillary pathway, which can be associated with neurological or optic nerve issues. It is not very specific to glaucoma itself but can be seen in some types.

Normal IOP (Intraocular Pressure): This is a very important clue. Glaucoma is typically associated with high IOP, but normal tension glaucoma is a known exception.

Large and deep cup: This is a classic sign of glaucomatous optic neuropathy. The cup-to-disc ratio is increased as the optic nerve fibers are lost.

Paracentral scotoma: This is a typical visual field defect seen in glaucoma, particularly in early to moderate stages.

Let's analyze each option against these findings:

A: Normal tension glaucoma (correct)

This fits perfectly with the normal IOP.

Large and deep cup and paracentral scotoma are also consistent with glaucoma.

Sluggish pupillary reaction, while not specific, can be seen in optic neuropathies.

This is a strong contender.

B: Absolute glaucoma

Absolute glaucoma is the end stage of glaucoma. At this stage, vision is severely impaired or lost, and often there is pain and very high IOP. The question mentions progressive deterioration not end stage. And IOP is normal.

Absolute glaucoma is less likely due to normal IOP.

Sluggish pupillary reaction might be present due to complete optic nerve damage.

Not a good fit because of normal IOP and 'progressive deterioration' (implies not end-stage yet).

C: Primary angle closure glaucoma

Primary angle closure glaucoma typically presents with acute symptoms like severe pain, red eye, blurred vision, halos around lights and high IOP due to angle closure. While chronic angle closure can be less symptomatic and IOP may fluctuate, normal IOP is less typical, especially when cup is large and deep and visual field is affected.

Sluggish pupillary reaction could be present due to acute angle closure impact.

Paracentral scotoma can occur in glaucoma, but acute presentation is more typical for initial symptoms of PACG.

Less likely due to normal IOP and the description being more of chronic progressive loss rather than acute attack.

D: Neovascular glaucoma

Neovascular glaucoma is secondary glaucoma associated with other systemic conditions like diabetes, retinal vein occlusion, etc., leading to new blood vessel formation in the angle of the eye, blocking aqueous outflow and causing high IOP.

Normal IOP makes this diagnosis highly improbable.

Sluggish pupillary reaction might be present if underlying condition affects pupillary pathways,

but not directly related to neovascular glaucoma itself.

Cup and visual field defects can occur but are secondary to underlying cause and glaucoma progression.

Unlikely due to normal IOP and no mention of underlying systemic conditions predisposing to neovascularisation.

Step-by-step thinking process:

Identify the key symptoms and signs: Progressive vision deterioration, sluggish pupillary reaction, normal IOP, large and deep cup, paracentral scotoma.

Focus on the most important and contrasting sign: Normal IOP. This immediately points towards normal tension glaucoma.

Evaluate each option against all findings:

Normal tension glaucoma: Fits well with normal IOP, large cup, paracentral scotoma and progressive deterioration. Sluggish pupillary reaction is less specific but possible.

Absolute glaucoma: Contradicted by normal IOP and description as 'progressive deterioration' not end-stage.

Primary angle closure glaucoma: Contradicted by normal IOP and described as progressive, not acute presentation.

Neovascular glaucoma: Contradicted by normal IOP.

Confirm the best fit: Normal tension glaucoma is the most probable diagnosis because it aligns with all the presented clinical findings, especially the normal IOP in the context of glaucomatous optic neuropathy (large cup and visual field defect).

Select the correct option: Option A: Normal tension glaucoma.

**Output Response:**

The patient's presentation strongly suggests Normal Tension Glaucoma (NTG). Here's a detailed explanation:

Normal IOP: The key characteristic of NTG is glaucomatous optic neuropathy occurring despite intraocular pressure (IOP) being within the statistically normal range (typically <21 mmHg). The question explicitly states 'IOP is normal', making NTG a primary consideration.

Progressive Deterioration of Vision: This is a hallmark symptom of glaucoma, including NTG, as the optic nerve damage progresses gradually over time, leading to a slow decline in visual function.

Large and Deep Cup: Fundoscopy revealing a 'large and deep cup' is a classic sign of glaucomatous optic nerve damage. The cup-to-disc ratio is increased in glaucoma as the neural rim of the optic disc is progressively lost due to retinal ganglion cell death. A deep cup further indicates significant excavation of the optic disc.

Paracentral Scotoma: Visual field testing demonstrating a 'paracentral scotoma' is a typical visual field defect in glaucoma. Paracentral scotomas are early field defects often seen in glaucoma, including NTG, affecting the area close to, but not directly at, the point of fixation. They occur due to damage to specific nerve fiber bundles in the optic nerve.

Sluggish Pupillary Reaction: While not a definitive diagnostic criterion for NTG, a sluggish pupillary reaction in a 70-year-old can be associated with age-related pupillary changes or potentially mild optic nerve dysfunction. It is less specific than the other findings but is not inconsistent with glaucoma, particularly NTG.

Why other options are less likely:

B: Absolute glaucoma: Absolute glaucoma represents the end-stage of glaucoma, leading to complete blindness. While the patient has 'progressive deterioration', it's less likely to be absolute glaucoma as the presentation doesn't explicitly mention complete blindness, and the description focuses on progressive changes. Moreover, absolute glaucoma often presents with fixed and dilated pupils, not just sluggish reaction.   Normal IOP is also less typical of end-stage glaucoma unless it's a case where IOP was previously elevated and then normalized later due to ciliary body shutdown in advanced stages.

C: Primary angle closure glaucoma (PACG): PACG typically presents with acute symptoms like sudden eye pain, redness, blurred vision, halos around lights, and often significantly elevated IOP during acute attacks.   While chronic angle closure glaucoma exists, it is less common and usually presents with at least intermittently elevated IOP or a history of such.   Normal IOP and a gradual, progressive vision loss with paracentral scotoma are not typical of PACG. Sluggish pupillary reaction might be seen in acute PACG, but the overall clinical picture is not consistent.

D: Neovascular glaucoma: Neovascular glaucoma is a secondary glaucoma caused by new blood vessel formation in the iris and angle, often due to conditions like diabetic retinopathy or retinal vein occlusion. It typically presents with elevated IOP due to blockage of aqueous outflow by these new vessels. Fundoscopy in neovascular glaucoma would often reveal neovascularization of the iris (rubeosis iridis) and angle, which is not mentioned. Normal IOP is very unlikely in established neovascular glaucoma unless it's a very early or atypical presentation, which is not suggested by the rest of the clinical picture.

In summary, the combination of progressive vision loss, normal IOP, large and deep cup, and paracentral scotoma strongly points towards Normal Tension Glaucoma as the most probable diagnosis.